%% file: main.tex
\documentclass[journal]{new-aiaa}
\usepackage[utf8]{inputenc}
\usepackage{textcomp}

\usepackage{graphicx}
\usepackage{amsmath}
\usepackage[version=4]{mhchem}
\usepackage{siunitx}
\usepackage{longtable,tabularx}

\usepackage{amsfonts}
\usepackage{lipsum}  
\usepackage[dvipsnames]{xcolor}
\usepackage{threeparttable}
\usepackage{fancyvrb} 
\usepackage{lipsum}

\usepackage{hyperref} 
\hypersetup{
    pageanchor=true,
    plainpages=false,
    bookmarksnumbered,
    colorlinks = true,
    citecolor = gray,
    filecolor = magenta,
    linkcolor = gray,
    urlcolor = magenta,
}

\input{common}

\title{\framework: A Large-Scale Dataset and Benchmark for Airport Surface Movement Forecasting}

\author{
Ingrid Navarro\footnote{Equal contribution}
Pablo Ortega-Kral$^{\ast}$, 
Jay Patrikar, 
Haichuan Wang, Alonso Cano, Zelin Ye, Jong Hoon Park, \\ Sebastian Scherer${^{\ddagger}}$ and Jean Oh\footnote{Equal Advising}
}

\affil{Carnegie Mellon University, Pittsburgh PA, USA, 15213}

\begin{document}

\maketitle

\input{sections/0_abstract}
\input{sections/1_introduction}
\input{sections/2_related_work}

\input{sections/3_dataset}
\input{sections/5_benchmark}
\input{sections/4_method}
\input{sections/7_results}
\input{sections/8_conclusions}

\input{sections/11_acknowledgement}
\bibliography{main}
\clearpage
\newpage
\appendix
\section*{Supplementary Material} \label{sec:appendix}

\input{appendix/1_framework_tools}
\input{appendix/2_dataset}

\input{appendix/3_dataset_analysis}

\input{appendix/4_benchmark}
\input{appendix/5_model}

\end{document}

%% file: common.tex
\usepackage[utf8]{inputenc}  
\usepackage{caption}
\usepackage{subcaption}
\usepackage{graphicx}
\usepackage{amsmath}
\usepackage[version=4]{mhchem}
\usepackage{siunitx}
\usepackage{longtable,tabularx}
\usepackage{booktabs} 
\usepackage{cleveref}
\usepackage{fancyhdr} 
\usepackage{tcolorbox}
\usepackage{colortbl} 
\usepackage{multirow}
\usepackage{nicefrac}       
\usepackage{microtype}      
\usepackage[T1]{fontenc}    
\usepackage{textcomp}
\usepackage{amsfonts}       
\usepackage[dvipsnames]{xcolor}
\usepackage{threeparttable}
\usepackage{fancyvrb} 
\usepackage{float} 

\usepackage{listings}

\definecolor{jsonkey}{rgb}{0.56, 0.13, 0.56}
\definecolor{jsonstring}{rgb}{0.0, 0.5, 0.0}

\crefname{lstlisting}{Listing}{Listings}

\lstdefinelanguage{json}{
    basicstyle=\ttfamily\small,
    showstringspaces=false,
    breaklines=true,
    frame=single,
    backgroundcolor=\color{gray!10},
    literate=
     *{0}{{{\color{black}0}}}{1}
      {1}{{{\color{black}1}}}{1}
      {2}{{{\color{black}2}}}{1}
      {3}{{{\color{black}3}}}{1}
      {4}{{{\color{black}4}}}{1}
      {5}{{{\color{black}5}}}{1}
      {6}{{{\color{black}6}}}{1}
      {7}{{{\color{black}7}}}{1}
      {8}{{{\color{black}8}}}{1}
      {9}{{{\color{black}9}}}{1}
      {:}{{{\color{black}:}}}{1}
      {,}{{{\color{black},}}}{1}
      {"}{{{\color{black}"}}}{1},
    morecomment=[l]{//},
    morestring=[b]",
    stringstyle=\color{jsonstring},
    identifierstyle=\color{jsonkey}
}

\newif\ifshowauthorcomment
\showauthorcommenttrue

\newcommand{\idest}{i.e., }
\newcommand{\exempli}{e.g., }

\newcommand{\wrt}{w.r.t. }

\newcommand{\trajair}{{\tt TrajAir}}

\newcommand{\framework}{{\textcolor{Black!70}{Amelia}}}

\newcommand{\amelia}{{\textcolor{Black!70}{\texttt{Amelia}}}}
\newcommand{\ameliatf}{{\textcolor{Black!70}{\texttt{Amelia-TF}}}}
\newcommand{\ameliadataset}{{\textcolor{Black!70}{\texttt{Amelia-42}}}}
\newcommand{\ameliasubset}{{\textcolor{Black!70}{\texttt{Amelia42-Mini}}}}
\newcommand{\ameliabench}{{\textcolor{Black!70}{\texttt{Amelia10-Bench}}}}
\newcommand{\multiairport}{{\textcolor{Black!70}{\texttt{MultiAirport}}}}
\newcommand{\singleairport}{{\textcolor{Black!70}{\texttt{SingleAirport}}}}

\newcommand{\attribute}[1]{\textcolor{BlueViolet}{\texttt{#1}}}
\newcommand{\icao}[1]{{\textcolor{Black!70}{\texttt{#1}}}}

\newcommand{\panc}{{\textcolor{Black!70}{\texttt{PANC}}}}
\newcommand{\kbos}{{\textcolor{Black!70}{\texttt{KBOS}}}}
\newcommand{\kdca}{{\textcolor{Black!70}{\texttt{KDCA}}}}
\newcommand{\kewr}{{\textcolor{Black!70}{\texttt{KEWR}}}}
\newcommand{\kjfk}{{\textcolor{Black!70}{\texttt{KJFK}}}}
\newcommand{\klax}{{\textcolor{Black!70}{\texttt{KLAX}}}}
\newcommand{\kmdw}{{\textcolor{Black!70}{\texttt{KMDW}}}}
\newcommand{\kmsy}{{\textcolor{Black!70}{\texttt{KMSY}}}}
\newcommand{\ksea}{{\textcolor{Black!70}{\texttt{KSEA}}}}
\newcommand{\ksfo}{{\textcolor{Black!70}{\texttt{KSFO}}}}

\newcommand{\klga}{{\textcolor{Black!70}{\texttt{KLGA}}}}
\newcommand{\kord}{{\textcolor{Black!70}{\texttt{KORD}}}}

\newcommand{\seen}[1]{{\textcolor{Black!70}{\texttt{#1-Seen}}}}

\newcommand{\state}{{\mathbf x}}

\newcommand{\context}{{\mathbf c}}

\newcommand{\R}{{\mathbb R}}
\newcommand{\N}{{\mathbb N}}

\usepackage{booktabs}

\newcommand{\wip}[1]{\ifshowauthorcomment\textcolor{Magenta}{\textbf{Work in progress: }}\fi}

\newcommand{\Runway}{{\textcolor{Green}{Runway}}}
\newcommand{\Taxiway}{{\textcolor{RoyalBlue}{Taxiway}}}
\newcommand{\Holdline}{{\textcolor{Orange!80}{Hold-short Line}}}

\newcommand{\runway}{{\textcolor{Green}{runway}}}
\newcommand{\taxiway}{{\textcolor{RoyalBlue}{taxiway}}}
\newcommand{\holdline}{{\textcolor{Orange!80}{hold-short line}}}

\newcommand{\aircraft}{{\textcolor{Red}{Aircraft}}}
\newcommand{\unknown}{{\textcolor{LimeGreen}{Unknown}}}
\newcommand{\vehicle}{{\textcolor{RoyalBlue}{Vehicle}}}

\usepackage{xargs} 
\usepackage[colorinlistoftodos,prependcaption,textsize=tiny]{todonotes}
\newcommandx{\unsure}[2][1=]{\todo[linecolor=red,backgroundcolor=red!25,bordercolor=red,#1]{#2}}
\newcommandx{\change}[2][1=]{\todo[linecolor=blue,backgroundcolor=blue!25,bordercolor=blue,#1]{#2}}
\newcommandx{\info}[2][1=]{\todo[linecolor=OliveGreen,backgroundcolor=OliveGreen!25,bordercolor=OliveGreen,#1]{#2}}
\newcommandx{\improvement}[2][1=]{\todo[linecolor=Plum,backgroundcolor=Plum!25,bordercolor=Plum,#1]{#2}}
\newcommandx{\thiswillnotshow}[2][1=]{\todo[disable,#1]{#2}}
 
\def\HiLiYellow{\leavevmode\rlap{\hbox to \hsize{\color{yellow!20}\leaders\hrule height .8\baselineskip depth .4ex\hfill}}}
  
\usepackage{cleveref}

%% file: sections/0_abstract.tex
\begin{figure}[h!]
    \vspace{-0.8cm}
    \centering%
    {\includegraphics[width=0.98\textwidth,trim={0cm 0cm -0cm -0cm},clip]{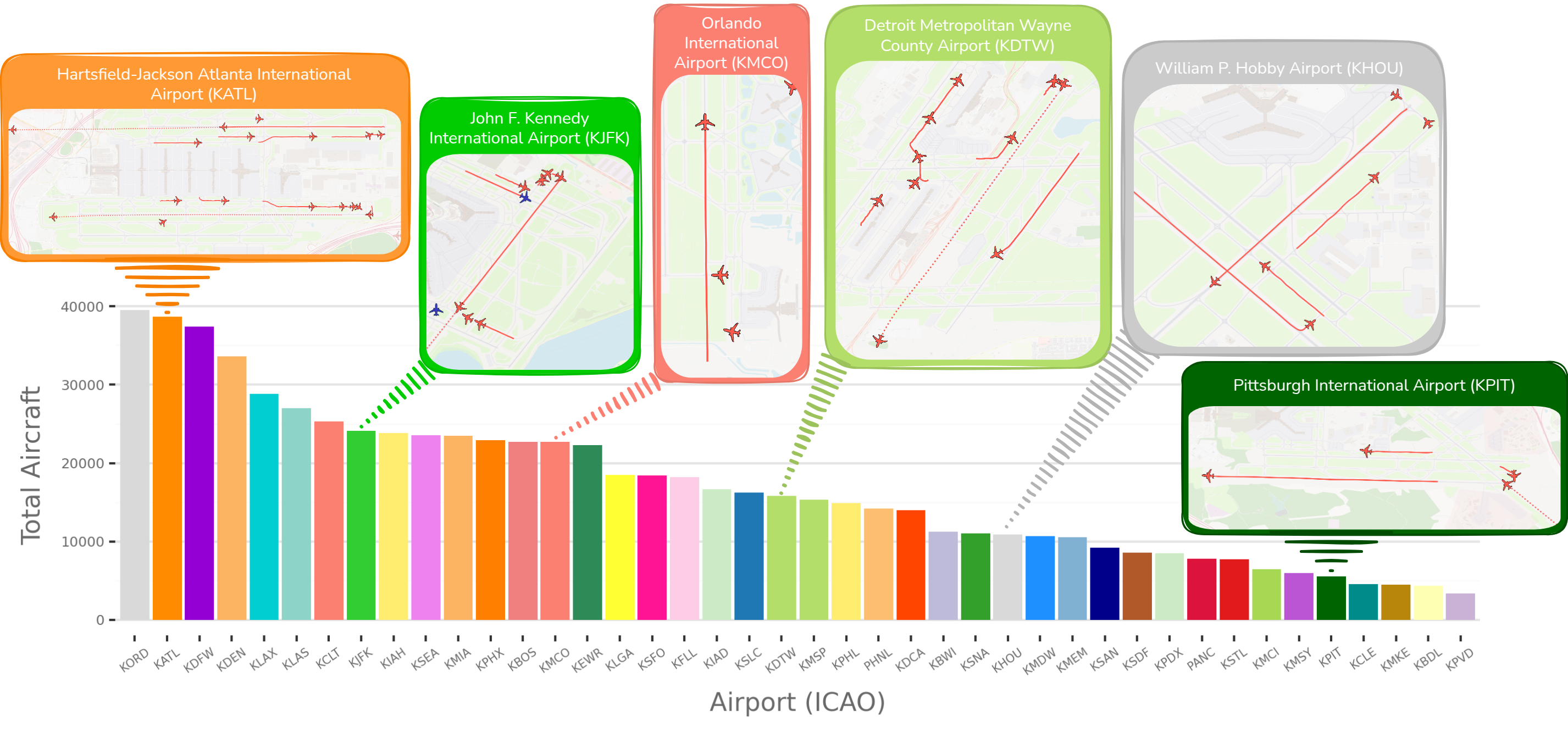}}
    \caption{The \ameliadataset~Dataset. Total aircraft count per airport over a span of 15 days. We also include examples of airport scenes across multiple airports, showing diverse behaviors and interactions.}
    \label{fig:teaser}
\end{figure}%

\begin{abstract}
Demand for air travel is rising, straining existing aviation infrastructure. In the US, more than 90\% of airport control towers are understaffed, falling short of FAA and union standards. This, in part, has contributed to an uptick in near-misses and safety-critical events, highlighting the need for advancements in air traffic management technologies to ensure safe and efficient operations. Data-driven predictive models for terminal airspace show potential to address these challenges; however, the lack of large-scale surface movement datasets in the public domain has hindered the development of scalable and generalizable approaches. To address this, we introduce \ameliadataset, a first-of-its-kind large collection of raw airport surface movements reports streamed through the FAA's System Wide Information Management (SWIM) Program, comprising over two years of trajectory data ($\sim$9.19TB) across 42 US airports. We open-source tools to process this data into clean tabular position reports. We release \ameliasubset, a 15-day sample per airport, fully processed data on HuggingFace for ease of use. We also present a trajectory forecasting benchmark consisting of \ameliabench, an accessible experiment family using 292 days from 10 airports, as well as \ameliatf, a transformer-based baseline for multi-agent trajectory forecasting. All resources are available at our website: \href{https://ameliacmu.github.io}{\texttt{ameliacmu.github.io}} and \href{https://huggingface.co/AmeliaCMU}{\texttt{huggingface.co/AmeliaCMU}}.
\end{abstract}

%% file: sections/1_introduction.tex
\section{Introduction} \label{sec:intro}

In early 2023, the U.S. National Airspace System (NAS) recorded more aircraft close-calls than in the previous five years combined~\cite{daugherty2023we}. Poorly managed airport operations pose serious risks. For instance, in recent events, the world has witnessed major incidents that have caused significant damage and, more importantly, resulted in the loss of lives~\cite{leussink2024japan, helmuth2025what, ruberg2025one}. While part of these alarming trends stem from deficiencies within Air Traffic Control (ATC) facilities~\cite{EmberSteel2023, Krolik2025}, growing airspace demand~\cite{iata2022air} and record-setting travel volumes~\cite{aratoni2023tsa} are further straining the existing infrastructure. Moreover, the emergence of Advanced Aerial Mobility (AAM) further highlights the need for automated, data-driven decision-support systems~\cite{degas2022survey} to safely integrate manned and unmanned aircraft. These challenges underscore the urgency of advancing air traffic management and aviation safety standards to mitigate potential failures and prevent catastrophic accidents.

Data-driven human-aware predictive approaches have shown great success in domains like autonomous driving~\cite{sun2020scalability, fang2019argoverse, caesar2020nuscenes, rudenko2020humanmotion} and crowd navigation~\cite{pellegrini2009eth, lerner2007ucy, rudenko2020humanmotion}, and hold significant potential for addressing \textbf{safety and efficiency} challenges in aviation. For example, they could be utilized by aircraft and management facilities to enhance situational awareness through onboard sensors for intelligent navigation~\cite{navarro2023sorts, patrikar2022challenges} and automated go-around decisions to prevent near misses and collisions~\cite{muthali2023multi,sui2023conflict, kong2023identifying, memarzadeh2023anomaly}. They could also improve efficiency~\cite{liu2014predictability}, for instance, by estimating taxi-out and departure times to coordinate operations~\cite{lee2016taxi, wang2023data}, de-icing schedules~\cite{alsalous2023deicing}, and time-of-arrival estimates~\cite{gui2021data}. However, the lack of unified frameworks and large-scale open datasets has limited the \textbf{scalability and generalizability} of existing solutions~\cite{li2019reviewing}. Most prior works rely on small datasets~\cite{navarro2023sorts, zhang2022airport, park2023influential} or simple prediction models that do not account for complex interactions and modalities of information~\cite{ranson2011faa}.

To address this gap, this paper introduces the \amelia~framework featuring four key contributions:

First, we present \ameliadataset, a large-scale dataset focused on airport surface operations, collected via the Federal Aviation Administration's (FAA) System Wide Information Management (SWIM) Program. Our dataset includes over 9.19TB of compressed surface movement trajectory data from 42 U.S. airports, starting from December 1, 2022, up to today. Alongside the SWIM data, we collect, post-process, and release airport surface maps as lightweight, semantically marked graphs. Then, for ease of accessibility, we release \ameliasubset, a clean, tabular 15-day per-airport (\idest~630 total days) sample via HuggingFace
, along with tools for downloading and processing additional data as needed
. To our knowledge, \ameliadataset~is the largest open dataset of its kind and is intended for use by researchers working on airport surface operations, motion forecasting, and beyond.

Second, we introduce \ameliabench, a trajectory forecasting benchmark designed to support the research and development of large predictive models for aviation, as well as scalable and generalizable learning strategies for motion forecasting. The benchmark covers over 290 days of data from 10 different airports
and includes a family of experiments across two settings: \singleairport~and \multiairport. The \singleairport~setting focuses on studying \textit{in-domain} performance, \idest training and testing model performance one airport at a time, while \multiairport~setting focuses on studying \textit{cross-domain} generalization capabilities of predictive models. 

Third, to provide baseline results across the proposed benchmark, we contribute \ameliatf
, a large transformer-based trajectory forecasting model, designed for predicting aircraft's future movements by characterizing aircraft's temporal information, and agent-to-agent and agent-to-map relationships. \ameliatf~leverages a novel automated scene representation strategy that abstracts and prioritizes scene information to improve model generalization. Using standard metrics for trajectory prediction, we provide both short- and long-horizon results across \ameliabench.

Finally, along with our dataset, we also release our model, model weights, and all tools used for collection and processing to promote large-scale open aviation research at: \href{https://ameliacmu.github.io}{\texttt{ameliacmu.github.io}} and \href{https://huggingface.co/AmeliaCMU}{\texttt{huggingface.co/AmeliaCMU}}

\footnote{\textbf{Note to the reader:} This manuscript extends our previous work \cite{navarro2024ameliatf}, which constitutes a non-archival publication. We briefly summarize the differences between the two versions: (1) We now focus on the improved access to the Amelia dataset by centralizing two processed versions: Amelia-42-mini and Amelia-10, both of which are now easily accessible on HuggingFace. (2) We provide more in-depth details about the proposed trajectory forecasting benchmark, and moved the model details to the appendix. (3) We extend our data analysis to provide motion profiles for all airports, heatmaps of dataset coverage, enhanced metadata reporting for the processed subsets, and further analysis of the timespans, most of which can be found in the appendix.}


%% file: sections/2_related_work.tex
\section{Related Work} \label{sec:relwork}

\paragraph{Trajectory Forecasting Datasets in Aviation.} In contrast to the autonomous driving (AD) domain, where large-scale datasets exist~\cite{sun2020scalability, fang2019argoverse, caesar2020nuscenes, rudenko2020humanmotion}, the domain of aviation has not been studied as extensively due in part to the lack of datasets. 
Previous works \cite{li2019reviewing} have criticized the dominance of proprietary, non-public data in aviation research, finding that 68\% of the works reviewed utilized solely proprietary data. 
In response, \trajair~\cite{patrikar2022predicting} was introduced as a dataset in the public domain for motion prediction in general aviation. It aimed to learn how pilots interact in non-towered airports but consisted only of trajectory information and focused only on a single airport. 
Accordingly, in response to the lack of curated large-scale publicly accessible datasets in this domain,  we release the \ameliadataset~dataset in its raw format, along with two curated subsets, \ameliasubset~and \ameliabench, for ease of accessibility to researchers in the forecasting domain. 

To the best of our knowledge, \ameliadataset~is the largest dataset of its kind and is geared towards use by researchers even beyond those interested in airport motion forecasting.    

\paragraph{Trajectory Forecasting Models in Aviation.} Trajectory forecasting is a well established field in autonomous driving \cite{gao2020vectornet, nayakanti2023wayformer, cui2023gorela, shi2022motion, yuan2021agentformer, ngiam2021scene} and crowd navigation \cite{navarro2022socialpattern, alahi2016social,zhao2020noticing, rudenko2020humanmotion, yuan2021agentformer}. Motivated by the growing challenges of coordinating air traffic, an emerging research area focuses on employing learning-based trajectory forecasting in the aviation domain. The bulk of prior art \cite{guo_flightbert_2023, guo_flightbert_2024, zhang_flight_2023} targets short-horizon in-air prediction. FlightBERT \cite{guo_flightbert_2023} is one of the first in this line of research. It binarizes each trajectory point and feeds the resulting bit-strings to a transformer, framing forecasting as a multi-binary classification problem. Closest to our setting is \cite{zhang2022airport}, where the authors analyze four hours' worth of movement at two airports, identifying the 2D position of each agent; and \cite{yin2023context}, who use one-week ADS-B data collected at Singapore Changi Airport. Neither dataset is in the public domain, and both studies are limited to a single airport, restricting generalization.

Our work differs significantly in scale and scope, as we wish to forecast trajectories for airport ground operations across multiple airport topologies; a setting that current datasets and methods have not fully addressed. Single-agent models such as FlightBERT lack mechanisms to capture inter-agent interactions, as well as failing to capture contextual information that conditions different behaviors. In addition, prior works' absolute trajectory representations can limit transfer between airports, whereas our multi-airport dataset necessitates efficient encoding for better large-scale training.

%% file: sections/3_dataset.tex
\begin{figure*}[t!]
    \centering
    \includegraphics[width=0.97\textwidth]{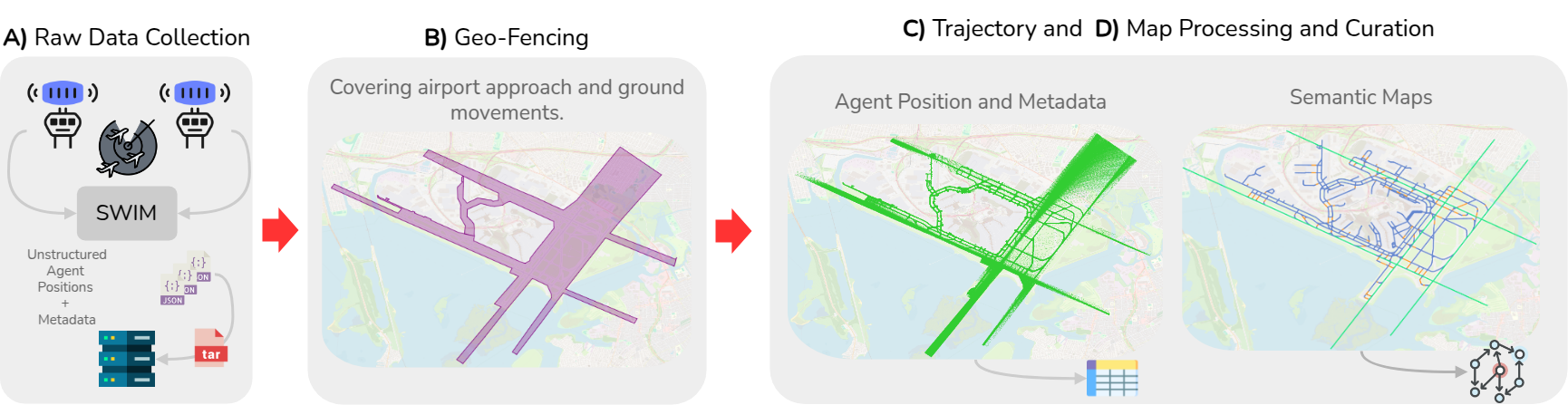}
    \caption{The \amelia~data pipeline. \textbf{A)} Raw position reports from the FAA's SWIM Terminal Data Distribution System are continuously logged and released as \ameliadataset, from December 2nd, 2022, to the present. \textbf{B)} Airport-specific geofences are defined to delimit movement areas as well as take-off and landing extensions to runways. \textbf{C)} Data within the geo-fence is processed into clean tabular 1-Hz position reports. \textbf{D)} As additional context, semantic routing graphs are created for each airport.}
    \label{fig:dataset_overview}
\end{figure*}

\section{Dataset Collection and Processing} \label{sec:dataset}

\textbf{Surface area operations} encompass landings, departures, and ground-vehicle activity within the airport movement zone. The \amelia~ framework focuses on data-driven behavior prediction within this area. To support this, we build a large-scale dataset by logging every position report that enters the zone, tracking each agent until it exits or crosses the non-movement boundary. These raw logs are then processed into tabular trajectories with agent metadata suitable for training trajectory forecasting models. The resulting data format is compatible with existing dataloaders for motion forecasting~\cite{ngiam2021scene}. We describe our collection and processing methodologies for the trajectory data in \Cref{ssec:trajectory_data} and map data \Cref{ssec:map_data}, with further detail in \Cref{asec:dataset}. An overview of the data pipeline is shown in \Cref{fig:dataset_overview}.


\subsection{Trajectory Data} \label{ssec:trajectory_data}

\paragraph{Raw Data.} \ameliadataset~leverages FAA's SWIM\footnote{System Wide Information Management program} Terminal Data Distribution System (STDDs), which aggregates multiple terminal-area feeds into a single stream of position reports for aircraft and vehicles operating within a few miles of the airport, covering \textit{approach} and \textit{ground} movement events. Each report arrives as an XML message that we continuously stream and group into newline-delimited JSON files, compressed for storage. Our data collection started on December 1st, 2022, and continues up to today. It currently represents \textbf{over 9.19~TB} of compressed position reports recorded and stored on our in-house server.

\paragraph{Processed Data.} Given that the raw position reports are irregularly sampled, have noise, and include many tracks outside of areas of movement, we further develop data pre-processing scripts following \cite{patrikar2022predicting}. Here, we focus on 42 airports in the US, for which we produce clean interpolated data in formats suitable for most modern ML-based trajectory forecasting dataloaders such as \cite{ivanovic2023trajdata}. We do so by defining a 3D geographical fence around the airport of interest, as shown in \Cref{fig:dataset_overview}, filtering the position reports that fall inside it and within 2000~ft above ground level. We also extract relevant metadata from each position report, which we then interpolate and resample at 1-Hz. We produce per-hour,per-airport tabular files in CSV format containing the information shown in  \Cref{tab:processed_data}. 

We open-source a toolkit for downloading and processing trajectory data from desired timeframes at \\
\href{https://github.com/AmeliaCMU/AmeliaSWIM}{github.com/AmeliaCMU/AmeliaSWIM}.

\input{tables/data_fields}

\subsection{Map Data} \label{ssec:map_data}

We use OpenStreetMap (OSM)~\cite{map2014open} and OSMNx~\cite{boeing2017osmnx} to create surface maps for the airports in \ameliadataset. Raw OSM airport maps are dense routing graphs containing centerline information of the movement areas, pavement boundaries, and semantic information, \exempli taxiways, runways, hold-short lines, exit lines, aprons, etc. Node elements within the graph contain geographical coordinates (latitude, longitude) and local Cartesian coordinates, $(x, y)$, expressed \wrt an arbitrary origin. 

We develop an automatic graph generation pipeline described in \Cref{assec:graph_pipeline} that sanitizes raw OSM maps into compact vectorized semantic graphs, following \cite{gao2020vectornet}. The provided maps are stored as a collection of vectors representing the edges pertaining only to the centerline of movement areas $G = \{v_0, v_1, \dots, v_E\}$. Each vector contains information about the start and endpoint, both in terms of GPS coordinates and local Cartesian Coordinates; a categorical value representing the class of the zone the edge belongs to; and a numeric identifier, see \Cref{tab:graph_attributes} for a breakdown. 

\input{tables/graph_attributes}

\section{Dataset Analysis} \label{ssec:subset}

To facilitate usability and accessibility of our dataset, we release a 630-day subset of processed data as \ameliasubset~at HuggingFace\footnote{Access the \ameliasubset~subset at: \href{https://huggingface.co/datasets/AmeliaCMU/Amelia42-Mini}{huggingface.co/datasets/AmeliaCMU/Amelia42-Mini}}. The subset covers 42 airports, randomly sampling 15 days per airport from the full dataset to capture diverse seasons and operational settings. In total, \ameliasubset~contains $\sim$8.43\textbf{M} unique agents, covering $\sim$1.10\textbf{B} position reports, as well as processed maps for each airport.

Below, we provide a trajectory and map analysis of the subset and refer the reader to \Cref{asec:subset} for a more in-depth, per-airport breakdown.

\begin{figure} [!ht]
    \centering
    \includegraphics[width=0.98\linewidth]{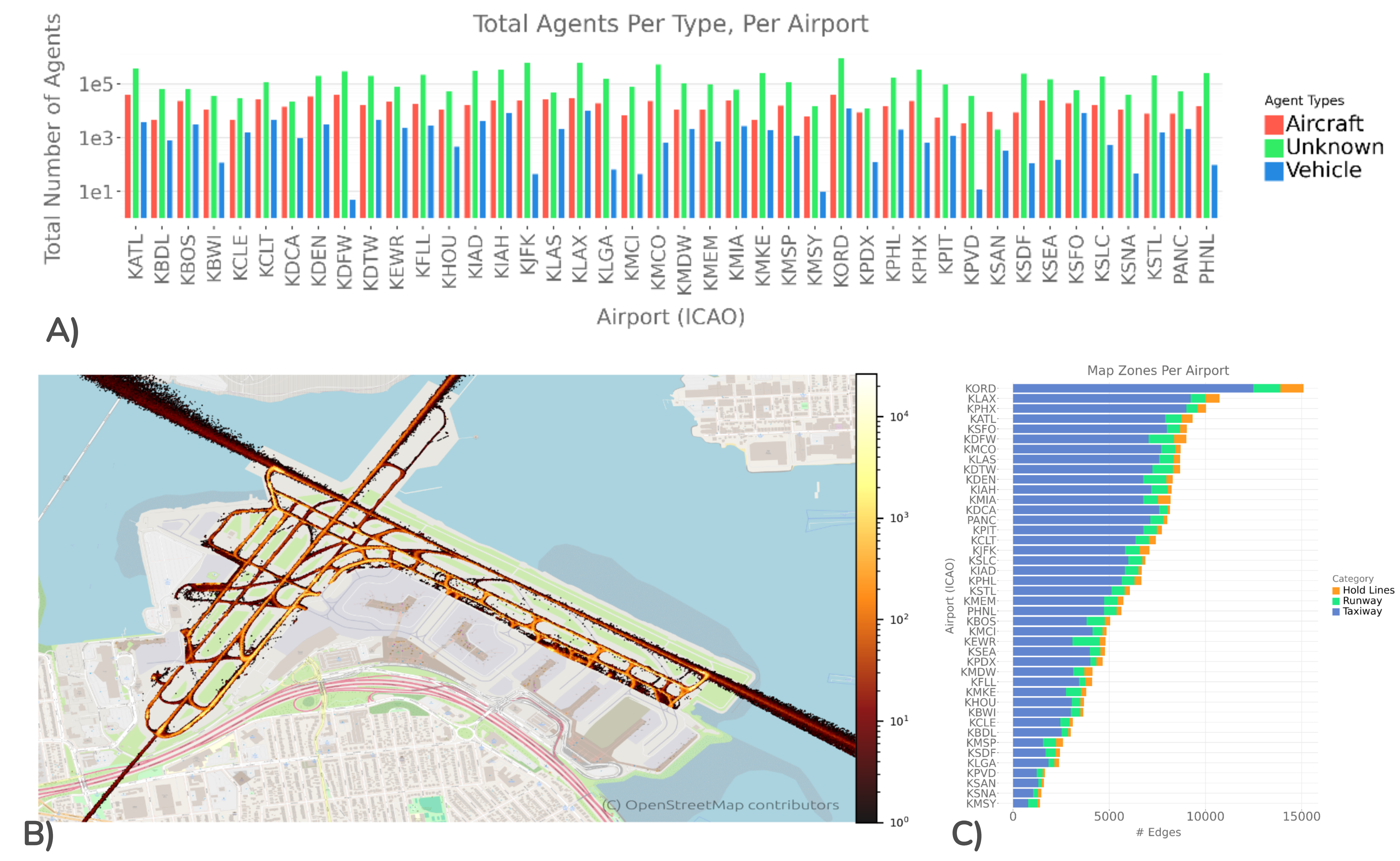}
    \caption{Analysis of the \ameliasubset~subset. \textbf{A)} Shows the total number of agents, per type (\aircraft, \unknown, \vehicle). \textbf{B)} Shows an airport's heatmap, representing the activity frequency per region in the airport (darker, low frequency; brighter, high frequency). \textbf{C)} Shows the distribution of zone types (\Taxiway, \Runway, \Holdline) per airport's map. } 
    \label{fig:dataset_analysis}
\end{figure}

\paragraph{Trajectory Data Analysis.} \Cref{fig:teaser} shows the total number of aircraft observed at each selected airport, identified by their International Civil Aviation Organization (ICAO) code. This highlights the variation in airport size and includes examples of diverse interaction scenarios. \Cref{fig:dataset_analysis}.\textbf{A)} presents the agent type breakdown per airport. Overall, the \ameliasubset~boasts over 708.88\textbf{k} aircraft, 90.14\textbf{k} vehicles, and 7.63\textbf{M} unknown agent types. As shown, the majority of the processed agents fall under the \textit{unknown} category. It’s important to note that this is not a result of our data processing pipeline, but rather a limitation inherent to the SWIM STTDS service and infrastructure utilized for data collection. Nonetheless, even if removing the \textit{unknown} agent types from our dataset, the scale and quality of the data are still substantial, and we believe that future work could explore re-labeling strategies to further enhance the quality of the data.

\Cref{fig:dataset_analysis}.\textbf{B)} illustrates the spatial data coverage for the La Guardia Airport (\klga), as a heatmap showing the activity frequency, bright spots represent concurred regions and capture critical section such as intersections and hold-lines; this showcases that our subset can capture operation hotspots relevant for behavior modeling. 

\paragraph{Map Data Analysis.} \Cref{fig:dataset_analysis}.\textbf{C)} summarizes the distribution of zone types per airport map. Each airport is shown as a stacked barplot presenting the number of edges per category in its corresponding semantic graph, \idest \Taxiway, \Runway, \Holdline. In all cases, the majority of the map information is comprised of taxiway edges, which represent the routes to and out of the runways, while critical regions like runways and hold lines are less predominant in each airport. The figure also shows that our map data has a diversity of map scales and densities. For instance, with airports like \kord~and \klax~with over 10k edges, and others like \kmsy~with ~1k edges. 

%% file: tables/data_fields.tex
\begin{table}[h!]
\centering
\caption{Fields with corresponding units and descriptions in the \ameliadataset~dataset. }
\resizebox{\textwidth}{!}{
\begin{tabular}{llllllll}
\toprule
\textbf{Field} & \textbf{Units}  & \textbf{Description}                & & &
\textbf{Field} & \textbf{Units} & \textbf{Description}                 \\ 
\cmidrule(l){1-3}\cmidrule(l){5-8}
\textcolor{BlueViolet}{\texttt{Frame}}          & \#             & Timestamp                           & & &
\textcolor{BlueViolet}{\texttt{Altitude}}       & feet           & Agent Altitude (Mean Sea Level)      \\ 
\textcolor{BlueViolet}{\texttt{ID}}             & \#             & STDDS Agent ID                      & & &
\textcolor{BlueViolet}{\texttt{Range}}          & km             & Distance from airport datum          \\ 
\textcolor{BlueViolet}{\texttt{Speed}}          & knots          & Agent Speed                         & & &
\textcolor{BlueViolet}{\texttt{Bearing}}        & rads           & Bearing Angle \wrt North             \\ 
\textcolor{BlueViolet}{\texttt{Heading}}        & degrees        & Agent Heading                       & & &
\textcolor{BlueViolet}{\texttt{Type}}           & int            & Agent Type {[}0:{\color{Red}A/C}, 1:{\color{RoyalBlue}Veh}, 2:{\color{green}Unk}{]} \\ 
\textcolor{BlueViolet}{\texttt{Lat}}            & decimal degs   & Latitude of the agent                  & & &
\textcolor{BlueViolet}{\texttt{x}}             & km             & Local X Cartesian Position           \\ 
\textcolor{BlueViolet}{\texttt{Lon}}            & decimal degs   & Longitude of the agent                 & & &
\textcolor{BlueViolet}{\texttt{y}}              & km             & Local Y Cartesian Position           \\ 
\textcolor{BlueViolet}{\texttt{Interp}}         & boolean        & Interpolated data point flag           & & &
                        &                &            \\ 
\bottomrule
\end{tabular}
}
\label{tab:processed_data}
\vspace{-0.3cm}
\end{table}

%% file: tables/graph_attributes.tex
\begin{table}[!ht]
\centering
\caption{Attributes for each of the edges in the vectorized graph.}
\resizebox{\textwidth}{!}{
\begin{tabular}{@{}lll@{}}
\toprule
\multicolumn{1}{c}{Field} & \multicolumn{1}{c}{Contents} & \multicolumn{1}{c}{Description}                                                                 \\ \midrule
\attribute{Start}                     & (latitude, longitude, y, x)  & Starting coordinates for the graph edge. We include both GPS coordinates as well as Cartesian coordinates. \\
\attribute{End}                       & (latitude, longitude, y, x)  & End coordinates for the graph edge.                                                             \\
\attribute{Semantic ID}               & Integer                      & Map Zone Type: {[}1: \holdline, 2: other, 3: \runway, 4: \taxiway{]}                        \\
\attribute{ID}                        & Integer                      & Unique Numeric Identifier                                                                       \\ \bottomrule
\end{tabular}
}
\label{tab:graph_attributes}
\end{table}

%% file: sections/5_benchmark.tex
\begin{figure*}[th!]
    \centering
    \includegraphics[width=0.98\textwidth]{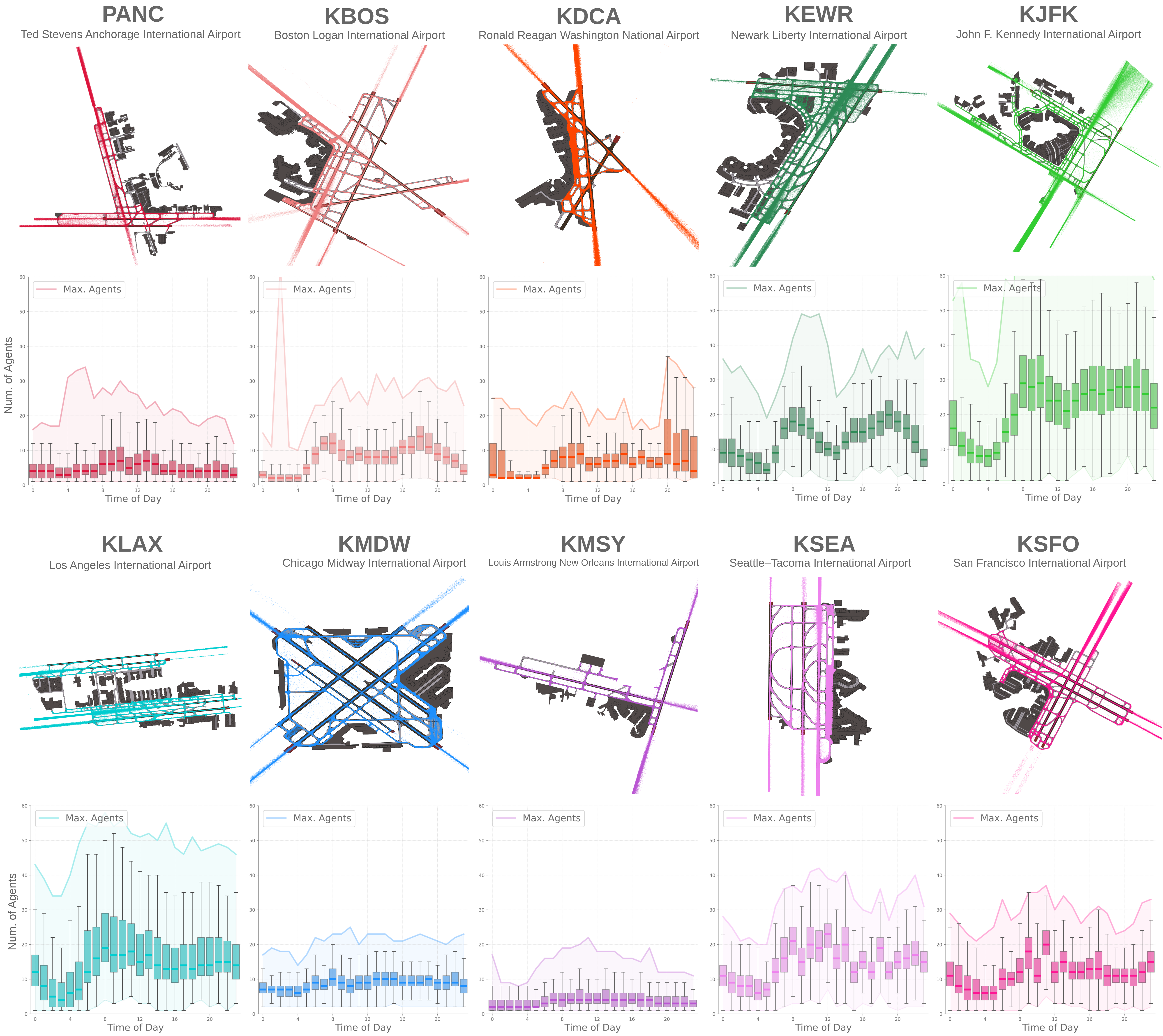}
    \caption{Analysis of the \ameliabench~data. For each airport, we overlay a month's worth of processed trajectories spanning beyond runway limits to capture landing and take-off rolls. We also show each airport's crowdedness per hour of day.}
    \label{fig:benchmark_data}
\end{figure*}

\section{A Trajectory Forecasting Benchmark} \label{sec:bech}

The \ameliabench~is a scenario benchmark for trajectory forecasting in crowded airport environments. The goal of trajectory forecasting is to predict the future scene given the previous observation, i.e., estimating the conditional probability distribution of the future scene, given the observations from the past time steps. 
For this task, we propose a scene representation that is long enough to capture long-horizon and diverse maneuvers in airspace.

\subsection{Task Setup} \label{ssec:setup}

\paragraph{Scenario Representation.} We split the data into $T=60$ seconds scenes to capture long-horizon interactions, where the first $H=10$ seconds are the historical segment and the remaining $F=50$ seconds are for prediction to encourage long-horizon and preemptive reasoning from minimal historical information.
To split the data into a \textit{train/val} and \textit{test} sets, we use a \textit{day}-based splitting strategy where 80\% of each airport's collected days are used for training/validation and the remaining 20\%, for testing.  


\paragraph{Benchmark Data.} For benchmarking, we limit to a 10-airport subset of the \ameliadataset~spanning 292 days to keep the size of data under 50~GB so that the task is computationally accessible, while still providing more than 20M diverse scenarios. The tabular subset is accessible at HuggingFace and we release the tools to process the data into scenarios for trajectory forecasting\footnote{Access the \ameliabench~subset at: \href{https://huggingface.co/datasets/AmeliaCMU/Amelia-10}{huggingface.co/datasets/AmeliaCMU/Amelia-10}, and the scenario processing tool at: \href{https://github.com/AmeliaCMU/AmeliaScenes}{github.com/AmeliaCMU/AmeliaScenes}}

The airports are chosen to ensure diversity in map topologies and traffic density as depicted in~\Cref{fig:benchmark_data}, but also based upon prior incident history~\cite{EmberSteel2023}. 
We provide a comprehensive analysis showing the behavioral and environmental diversity of the data utilized for this study in \Cref{asec:benchmark_extra}.

The benchmark provides rich data that can support various experiments in airspace trajectory forecasting. As shown in~\Cref{fig:benchmark_data}, each airport presents a large number of specific challenges, which form a unique domain. This benchmark can be used for in-domain training of specialized models, e.g., predicting a landing at the Seattle-Tacoma Airport after training on that setting. At the same time, this benchmark can also be used for cross-domain generalization to develop robust models that can handle uncertainty in unseen airports. We show both examples in~\Cref{subsec:exp}.

\subsection{Experiment Family}\label{subsec:exp}

For the purpose of our experiments for \textit{in-domain} and \textit{cross-domain} evaluation, we propose two main challenges. The first is \singleairport, where we focus on assessing \textit{in-domain} trajectory forecasting performance by training and testing our model's performance one airport at a time. This setting, therefore, consists of 10 experiments, one per environment, and serves to encourage the development of specialized models. 

The second challenge is \multiairport, which consists of various experiments to assess \textit{cross-domain} generalization of predictive models as we increasingly constrain the number of training environments. The sweep consists of 6 experiments: \seen{10}, \seen{7}, \seen{4}, \seen{3}, \seen{2}, and \seen{1}, each named according to the number of training (\textit{seen}) airports\footnote{The experiments with a greater number of \textit{seen} airports are supersets of those with a smaller number.}. This proposed challenge aims to focus on developing representation learning strategies that enable zero-shot generalization. Therefore, it aims to minimize the performance drop between \textit{in-domain} and \textit{cross-domain} and, in doing so, identify a practical compromise between training-set scale and cross-airport generalizability, ultimately bridging the gap between \singleairport~and \multiairport. In \Cref{assec:bench_exp_family}, we further detail the airport selection criteria for this challenge. 

%% file: sections/4_method.tex
\begin{figure*}[t!]
    \centering
    \includegraphics[width=0.98\textwidth]{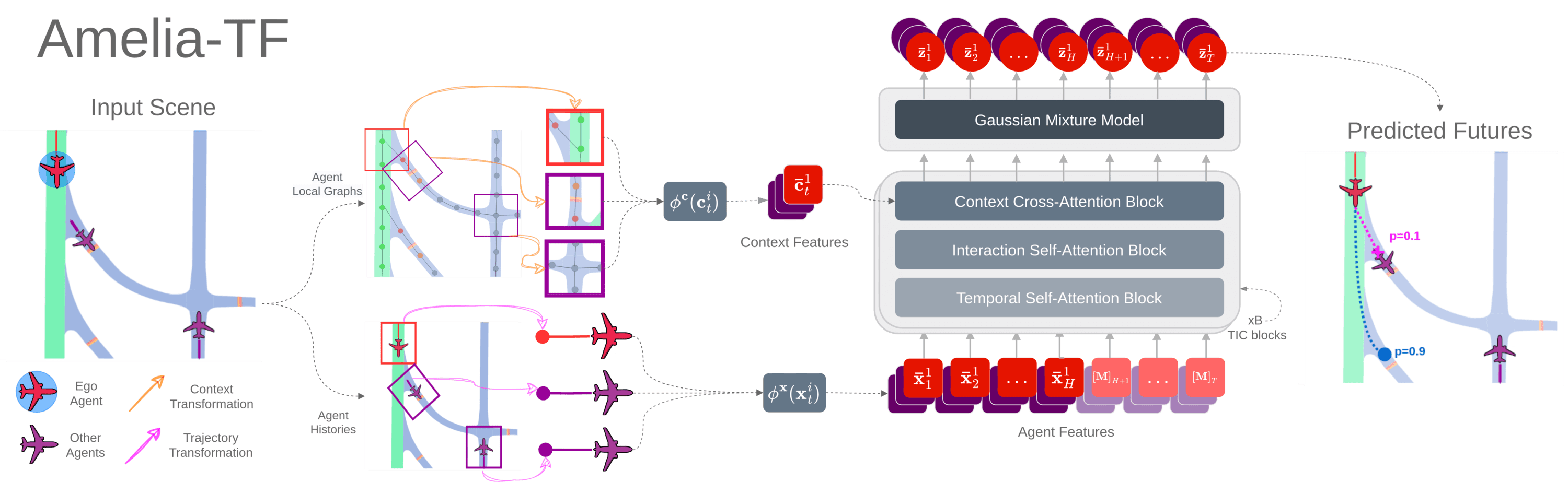}
    \caption{An Overview of \ameliatf. 
    }
    \label{fig:model}%
    \vspace{-0.2cm}%
\end{figure*}

\section{A Trajectory Forecasting Baseline} \label{sec:model}

To provide baseline results for the \ameliabench~benchmark, we introduce \ameliatf, a trajectory forecasting model that aims to characterize \textit{safety-relevant} airport surface movements. We build upon the intuitions from prominent state-of-the-art models to design a model capable of learning strong feature representations \cite{ngiam2021scene, shi2022motion}, as well as automated methods for selecting relevant scenes and agents to characterize \cite{stoler2023safeshift}. An overview of the model is shown in \Cref{fig:model}, and briefly described below. Full algorithmic details on safety-critical scene representation and the model implementation are described in \Cref{assec:model_scene} and \Cref{assec:implementation}, respectively. 

\subsection{Model Architecture} \label{ssec:model}

\paragraph{Scene Encoder.} To encode the scene, we follow a transformer-based architecture that uses a factorized attention scheme for efficient computations as in \cite{ngiam2021scene, ho2019axial}. The model maintains a $K \times F \times D$ representation across all transformer layers while interleaving two types of self-attention layers: a \textit{temporal} and an \textit{interaction} one, as well as a \textit{context} cross-attention layer as shown in \Cref{fig:model}. 
Our scene encoder consists of multiple blocks combining these layers sequentially to obtain a richer representation (See \Cref{assec:implementation}). 

\paragraph{Trajectory Decoder.} To model the distribution of possible futures for a scene, we adopt a Gaussian Mixture Model (GMM) as \cite{shi2022motion, chai2019multipath}, modeled as an MLP which takes the scene encoder's final output features and produces and decodes $F$ future timesteps over $M$ modes as $(\boldsymbol{\mu}, \boldsymbol{\Sigma}, \rho)^i_{m, t}$ representing the mean, covariance and likelihood over a state for an agent, mode, and timestep. 


\paragraph{Training Loss.} We formulate the following loss function, $\mathcal{L} = \mathcal{L}^{nll} + \mathcal{L}^{ce}$, where $\mathcal{L}^{nll}$ is a \textit{regression} component which back-propagates the average negative log-likelihood between the predictions and ground truth, and $\mathcal{L}^{ce}$ is a \textit{classification} component which considers the cross-entropy between the model's predicted confidences for each of the $M$ prediction heads and the one-hot encoding of the predicted future with the smallest error. 


%% file: sections/7_results.tex
\subsection{Baseline Results on the \ameliabench} \label{ssec:results}

We now study the performance of \ameliatf~on the \ameliabench. We thus report short-horizon results in \Cref{tab:multiairpot_results_t20} and long-horizon results in \Cref{tab:multiairpot_results_t50}. Each column shows the mADE/mFDE results for a given airport's test set, and the last column shows the average across all airports. For reference, the first row shows \singleairport~results, and each of the remaining rows represents a \multiairport~experiment. In \multiairport, the white and gray cells represent the seen (\textit{in-domain}) and unseen (\textit{cross-domain}) airports for the given experiment, respectively. We \textbf{bold} the best mADE/mFDE values across the experiment sweep and show in \textcolor{Green}{green} the values that were lower than the results obtained in \singleairport. 

\paragraph{Short-Horizon Trajectory Forecasting Results.} Our experiments show that generalization across unseen airports improves significantly as we increase the diversity and scale of seen environments. However, we also observe that our model generally exhibits strong performance even in more data-constrained experiments. We attribute this, in part, to our scene representation strategy, which aims to abstract and prioritize scene information more effectively. Nonetheless, a gap between experiments across the sweep still remains. The results show that the \seen{7} experiment achieves the lowest average mADE/mFDE from the sweep of experiments, closing the gap to \singleairport~to +1.60\%/-2.35\%. Furthermore, results across all airports as well the results for \ksea, \kdca, \panc~and \klax~as lower than those obtained in \singleairport.

\input{tables/generalization_t20}

\paragraph{Long-Horizon Trajectory Forecasting Results.} Our experiments also show that generalization across unseen airports improves significantly as we increase the diversity and scale of seen environments. However, the gap between experiments is further exacerbated in contrast to short-horizon predictions. The results show that the \seen{All} experiment achieves the lowest average mADE/mFDE from the full sweep of experiments, closing the gap to \singleairport~to -4.68\%/-4.83\%. Our model achieves better results for \panc, \klax, \kjfk~than in \singleairport.  

\input{tables/generalization_t50}

We present various scene examples with prediction results in \Cref{assec:results_forecasting} and further scenario analyses in \Cref{assec:results_scene}. 

%% file: tables/generalization_t20.tex
\begin{table*}[h!]
\centering
\caption{We show \ameliabench~results for a prediction horizon of \textbf{F=20}. }
\resizebox{0.98\textwidth}{!}{
\begin{tabular}{@{}lcccccccccccc@{}}
\toprule
    & \multicolumn{1}{c}{\textbf{\kmdw}} 
    & \multicolumn{1}{c}{\textbf{\kewr}} 
    & \multicolumn{1}{c}{\textbf{\kbos}} 
    & \multicolumn{1}{c}{\textbf{\ksfo}}
    & \multicolumn{1}{c}{\textbf{\ksea}}
    & \multicolumn{1}{c}{\textbf{\kdca}}
    & \multicolumn{1}{c}{\textbf{\panc}}
    & \multicolumn{1}{c}{\textbf{\klax}}
    & \multicolumn{1}{c}{\textbf{\kmsy}}
    & \multicolumn{1}{c}{\textbf{\kjfk}}
    & \multicolumn{1}{c}{\textbf{Average}} \\ 
\rowcolor[HTML]{FFFFFF} 
\textbf{Experiment} & mADE / mFDE & mADE / mFDE & mADE / mFDE & mADE / mFDE & mADE / mFDE & mADE / mFDE & mADE / mFDE  & mADE / mFDE & mADE / mFDE & mADE / mFDE & mADE / mFDE \\[0.2em]
\midrule
\textcolor{Black!70}{\texttt{Single-Airport}} 
    & 3.30 / 6.12
    & 6.61 / 12.92
    & 5.58 / 10.90
    & 5.06 / 9.82
    & 9.76 / 18.35
    & 4.74 / 9.22
    & 10.11 / 20.87
    & 11.36 / 20.63
    & 2.73 / 5.12
    & 4.58 / 9.52
    & 6.38 / 12.35 \\[0.2em]
\midrule
\seen{10}  						
    & 3.88 / 7.70
    & 7.87 / 15.80
    & 6.87 / 14.34
    & 6.09 / 12.64
    & \textcolor{Green}{9.03} / \textcolor{Green}{18.35}
    & 4.84 / 9.55
    & \textcolor{Green}{8.24} / \textcolor{Green}{16.75}
    & \textbf{\textcolor{Green}{8.80}} / \textbf{\textcolor{Green}{16.73}}	
    & \textbf{3.22} / \textbf{6.25}
    & \textbf{\textcolor{Green}{4.56}} / \textbf{9.64}
    &  \textcolor{Green}{6.34 (+0.70\%)} / 12.77 (-3.35\%) \\
\seen{7}							
    & 3.59 / 7.03
    & 7.30 / 14.54
    & 6.59 / 13.59
    & 5.73 / 11.65
    & \textcolor{Green}{\textbf{8.30}} / \textcolor{Green}{\textbf{16.71}}
    & \textbf{\textcolor{Green}{4.55}} / \textbf{\textcolor{Green}{8.82}}
    & \textbf{\textcolor{Green}{7.48}} / \textbf{\textcolor{Green}{14.94}}
    & \cellcolor[HTML]{EFEFEF} \textcolor{Green}{9.99} / \textcolor{Green}{19.35}
    & \cellcolor[HTML]{EFEFEF} 4.64 / 9.85
    & \cellcolor[HTML]{EFEFEF} 4.66 / 9.96
    & \textbf{\textcolor{Green}{6.28} (+1.60\%)} / \textbf{12.64 (-2.35\%)} \\ 
\seen{4}  							
    & 3.52 / 6.74
    & 7.26 / 14.33
    & 6.31 / 12.68
    & \textbf{5.66} / \textbf{11.33}
    & \cellcolor[HTML]{EFEFEF} 9.79 / 20.42
    & \cellcolor[HTML]{EFEFEF} 5.99 / 12.28
    & \cellcolor[HTML]{EFEFEF} \textcolor{Green}{9.22} / \textcolor{Green}{19.14}
    & \cellcolor[HTML]{EFEFEF} \textcolor{Green}{10.70} / 21.16
    & \cellcolor[HTML]{EFEFEF} 4.18 / 9.05
    & \cellcolor[HTML]{EFEFEF} 4.80 / 10.27
    & \ \ 6.74 (-5.32\%) / 13.74 (-10.15\%) \\
\seen{3} 									
    & \textbf{\textcolor{Green}{3.26}} / \textbf{6.59}
    & 7.25 / 14.20
    & \textbf{6.05} / \textbf{12.11}
    & \cellcolor[HTML]{EFEFEF} 7.25 / 15.50
    & \cellcolor[HTML]{EFEFEF} 9.90 / 20.95
    & \cellcolor[HTML]{EFEFEF} 6.16 / 12.74
    & \cellcolor[HTML]{EFEFEF} 9.53 / 20.26
    & \cellcolor[HTML]{EFEFEF} 10.99 / 21.86
    & \cellcolor[HTML]{EFEFEF} 4.33 / 9.29
    & \cellcolor[HTML]{EFEFEF} 4.96 / 10.54
    & \ \ 6.97 (-8.74\%) / 14.40 (-14.29\%) \\
\seen{2} 
    & 3.31 / 6.23
    & \textbf{6.92} / \textbf{13.45}
    & \cellcolor[HTML]{EFEFEF} 8.17	/ 17.63
    & \cellcolor[HTML]{EFEFEF} 7.49	/ 17.18 
    & \cellcolor[HTML]{EFEFEF} 10.57 / 22.86 
    & \cellcolor[HTML]{EFEFEF} 6.04	/ 12.47
    & \cellcolor[HTML]{EFEFEF} 10.61 / 23.00
    & \cellcolor[HTML]{EFEFEF}  12.78 / 26.34
    & \cellcolor[HTML]{EFEFEF}  4.49 / 9.64 
    & \cellcolor[HTML]{EFEFEF} 5.21 / 11.15
    &  \ 7.56 (-15.56\%)  / 15.99 (-22.81\%) \\ 
\seen{1}  
    & 3.30 / 6.12
    & \cellcolor[HTML]{EFEFEF} 13.76 / 30.91 
    & \cellcolor[HTML]{EFEFEF} 11.30 / 25.58 
    & \cellcolor[HTML]{EFEFEF} 9.68 / 21.73 
    & \cellcolor[HTML]{EFEFEF} 14.07 / 31.44
    & \cellcolor[HTML]{EFEFEF} 7.80 / 16.07 
    & \cellcolor[HTML]{EFEFEF} 15.00 / 33.75 
    & \cellcolor[HTML]{EFEFEF} 15.49 / 33.13 
    & \cellcolor[HTML]{EFEFEF}  7.43 / 17.11
    & \cellcolor[HTML]{EFEFEF}  7.40 / 16.77 
    &  10.52 (-39.33\%) / 23.26 (-46.92\%) \\  
\bottomrule
\end{tabular}
}
\label{tab:multiairpot_results_t20}
\end{table*}

%% file: tables/generalization_t50.tex
\begin{table*}[!ht]
\centering
\caption{We show \ameliabench~results for a prediction horizon of \textbf{F=50}, reported in meters. 
}
\resizebox{\textwidth}{!}{
\begin{tabular}{@{}lcccccccccccc@{}}
\toprule
    & \multicolumn{1}{c}{\textbf{\kmdw}} 
    & \multicolumn{1}{c}{\textbf{\kewr}} 
    & \multicolumn{1}{c}{\textbf{\kbos}} 
    & \multicolumn{1}{c}{\textbf{\ksfo}}
    & \multicolumn{1}{c}{\textbf{\ksea}}
    & \multicolumn{1}{c}{\textbf{\kdca}}
    & \multicolumn{1}{c}{\textbf{\panc}}
    & \multicolumn{1}{c}{\textbf{\klax}}
    & \multicolumn{1}{c}{\textbf{\kmsy}}
    & \multicolumn{1}{c}{\textbf{\kjfk}}
    & \multicolumn{1}{c}{\textbf{Average}} \\ 
\rowcolor[HTML]{FFFFFF} 
\textbf{Experiment} & mADE / mFDE & mADE / mFDE & mADE / mFDE & mADE / mFDE & mADE / mFDE & mADE / mFDE & mADE / mFDE  & mADE / mFDE & mADE / mFDE & mADE / mFDE & mADE / mFDE \\[0.2em]
\midrule
\textcolor{Black!70}{\texttt{Single-Airport}} 
    & 11.50	/ 28.80	
    & 23.68	/ 57.63	
    & 21.34	/ 53.76	
    & 17.05	/ 40.23	
    & 29.94	/ 65.82	
    & 16.42	/ 40.57	
    & 38.84	/ 101.89
    & 36.08	/ 88.25	
    & 9.89	/ 25.68	
    & 17.11	/ 41.19	
    & 22.18 / 54.39 \\[0.2em]
\midrule
\seen{All}  
    & 15.30 / 40.91 
    & 28.76 / 70.23
    & 28.52 / 73.57 
    & 22.63 / 53.78
    & 30.41 / 68.98
    & 17.86 / 44.88
    & \textcolor{Green}{30.19} / \textcolor{Green}{72.78}
    & \textbf{\textcolor{Green}{29.73}} / \textbf{\textcolor{Green}{72.46}}
    & \textbf{12.31} / \textbf{ 33.04}
    & \textbf{\textcolor{Green}{17.00}} / \textbf{\textcolor{Green}{40.92}}
    & \textbf{23.27} (-4.68\%) / \textbf{57.16} (-4.83\%) \\ 
\seen{7} 								
    & 14.35	/ 38.62
    & 26.83	/ 66.12
    & 27.57	/ 71.84
    & 20.81	/ 50.01
    & \textbf{\textcolor{Green}{28.29}} / \textbf{\textcolor{Green}{64.69}}
    & \textbf{\textcolor{Green}{16.40}} / \textbf{41.25}
    & \textbf{\textcolor{Green}{27.40}} / \textbf{\textcolor{Green}{66.74}}
    & \cellcolor[HTML]{EFEFEF} \textcolor{Green}{34.60} / \textcolor{Green}{85.90}
    & \cellcolor[HTML]{EFEFEF} 19.87 / 54.56
    & \cellcolor[HTML]{EFEFEF} 17.71 / 42.84
    & 23.38 (-5.12\%) / 58.26 (-6.63\%) \\ 
\seen{4}  												
    & 12.71	/ 31.64
    & 25.54	/ 59.18
    & 23.88	/ 58.45
    & \textbf{19.78} / \textbf{46.72}
    & \cellcolor[HTML]{EFEFEF} 33.36 / 73.86
    & \cellcolor[HTML]{EFEFEF} 22.40 / 54.04
    & \cellcolor[HTML]{EFEFEF} \textcolor{Green}{35.24} / \textcolor{Green}{83.90}
    & \cellcolor[HTML]{EFEFEF} 36.42 / \textcolor{Green}{85.11}
    & \cellcolor[HTML]{EFEFEF} 17.93 / 47.47
    & \cellcolor[HTML]{EFEFEF} 18.58 / 43.97
    &  24.58 (-9.76\%) / 58.43 (-6.92\%) \\
\seen{3} 		
    & 12.46	/ 31.81
    & 25.15	/ \textbf{\textcolor{Green}{57.29}}
    & \textbf{23.20} / \textbf{56.48}
    & \cellcolor[HTML]{EFEFEF} 28.33 / 67.40
    & \cellcolor[HTML]{EFEFEF} 35.70 / 81.02
    & \cellcolor[HTML]{EFEFEF} 24.32 / 59.55
    & \cellcolor[HTML]{EFEFEF} 38.85 / \textcolor{Green}{92.94}
    & \cellcolor[HTML]{EFEFEF} 39.02 / 91.96
    & \cellcolor[HTML]{EFEFEF} 19.53 / 51.52
    & \cellcolor[HTML]{EFEFEF} 19.37 / 47.06
    &  26.59 (-16.57\%) / 63.70 (-14.61\%) \\
\seen{2} 
    & 11.84	/ 29.89 
    & \textbf{24.62} / 59.04
    & \cellcolor[HTML]{EFEFEF} 34.89 / 88.52
    & \cellcolor[HTML]{EFEFEF} 31.69 / 78.28 
    & \cellcolor[HTML]{EFEFEF} 40.37 / 96.33 
    & \cellcolor[HTML]{EFEFEF} 24.00 / 62.72
    & \cellcolor[HTML]{EFEFEF} 43.30 / 107.04
    & \cellcolor[HTML]{EFEFEF} 48.91 / 122.31
    & \cellcolor[HTML]{EFEFEF} 19.29 / 51.33 
    & \cellcolor[HTML]{EFEFEF} 20.54 / 51.20
    & 29.94 (-25.92\%) / 74.67 (-27.15\%) \\ 
\seen{1}  
    & \textbf{11.50} / \textbf{28.80}
    & \cellcolor[HTML]{EFEFEF} 56.14 / 140.59
    & \cellcolor[HTML]{EFEFEF} 48.16 / 124.34 
    & \cellcolor[HTML]{EFEFEF} 40.54 / 103.97 
    & \cellcolor[HTML]{EFEFEF} 54.24 / 130.67
    & \cellcolor[HTML]{EFEFEF} 28.90 / 73.87
    & \cellcolor[HTML]{EFEFEF} 61.62 / 156.56
    & \cellcolor[HTML]{EFEFEF} 63.20 / 164.52 
    & \cellcolor[HTML]{EFEFEF}  30.59 / 80.23
    & \cellcolor[HTML]{EFEFEF}  29.51 / 73.97 
    & \ 42.44 (-47.73\%) / 107.75 (-49.52\%) \\  
\bottomrule
\end{tabular}
}
\label{tab:multiairpot_results_t50}
\end{table*}

%% file: sections/8_conclusions.tex
\section{Limitations \& Future Work} \label{ssec:future} 

While this work marks a significant advancement in facilitating large-scale open-source aviation research, it is not without its limitations. Here, we outline three interesting areas for future work in this field:


\paragraph{Managing Dataset Scale and Data Curation} \ameliadataset~ logs over 2 years of position reports across 42 airports in the US National Airspace, the sheer scale of this data makes processing and curating a significant challenge. Within the scope of this project, we release random subsets of data in an attempt to provide diverse settings while introducing minimal bias. However, future works could explore deriving different subsets from \ameliadataset~for specialized research lines; for instance, mining safety-critical incidents.

\paragraph{Tackling Generalizability and Long-Horizon Reasoning Challenges in Motion Forecasting.} \ameliatf~was designed and introduced as a baseline to encourage further research in motion forecasting in aviation. We believe that crafting representation learning strategies that facilitate domain adaptation and generalization to previously unseen situations offers a promising path for future exploration. Additionally, we see aviation as an ideal field for enhancing and refining the long-horizon reasoning capabilities of predictive models, as long-horizon prediction could facilitate the preemptive identification and remediation of critical scenarios.

\paragraph{Downstream Applications of Motion Forecasting in Aviation.} Predictive models for airport operations can be used for various tasks. For instance, \ameliatf~model can be combined with provably safe methods  \cite{muthali2023multi, nakamura2023online} to provide data-driven collision risk assessment and decision-support systems to mitigate near misses and critical scenarios. Other avenues and applications could include leveraging models like \ameliatf~to improve taxi-out time predictions, departure metering, and time-of-arrival calculations for more efficient coordination of airport surface operations.  

\section{Conclusions} \label{ssec:conclusion}

This work contributes \ameliadataset, a large-scale dataset for airport surface movement, which comprises trajectory information and graph-based representations of the airport maps. We validate our dataset and data processing pipelines, providing statistical analyses, visualizations, and insights for 42 major U.S. airports. To the best of our knowledge, this is the \textbf{largest dataset of its kind in the public domain}, specifically geared toward training next-generation data-driven predictive models for airport surface operations. To encourage further research, we open-source both our code and data.

We also introduce \ameliabench, a benchmark for trajectory forecasting in aviation, purposely designed to test the ability of architectures to leverage data for generalization to unseen settings. We contribute \ameliatf~as a baseline trajectory forecasting model, showcasing strong performance across various axes of analysis, exhibiting multi-future zero-shot generalization capabilities to previously unseen airports.

%% file: sections/11_acknowledgement.tex
\newpage
\section*{Funding Sources}

This work was supported by Boeing (award \#2022UIPA422). This work used Bridges-2 at Pittsburgh Supercomputing Center through allocation \verb|cis220039p| from the Advanced Cyberinfrastructure Coordination Ecosystem: Services \& Support (ACCESS) program, which is supported by National Science Foundation grants \#2138259, \#2138286, \#2138307, \#2137603, and \#2138296. This work was also supported by the Korean Ministry of Trade, Industry, and Energy (MOTIE; grant \#P0026022), and by the Korea Institute of Advancement of Technology (KIAT), through the International Cooperative R\&D program (\#P0019782): Embedded AI Based fully autonomous driving software and Maas technology development. 

\section*{Acknowledgments}

We thank Charles Erignac and Chad McFarland for their support and feedback throughout the project. 

%% file: appendix/1_framework_tools.tex
\section{\amelia~Framework Supplementary} \label[appendix]{asec:framework}

\begin{figure*}[!ht]
    \centering
    \includegraphics[width=\textwidth]{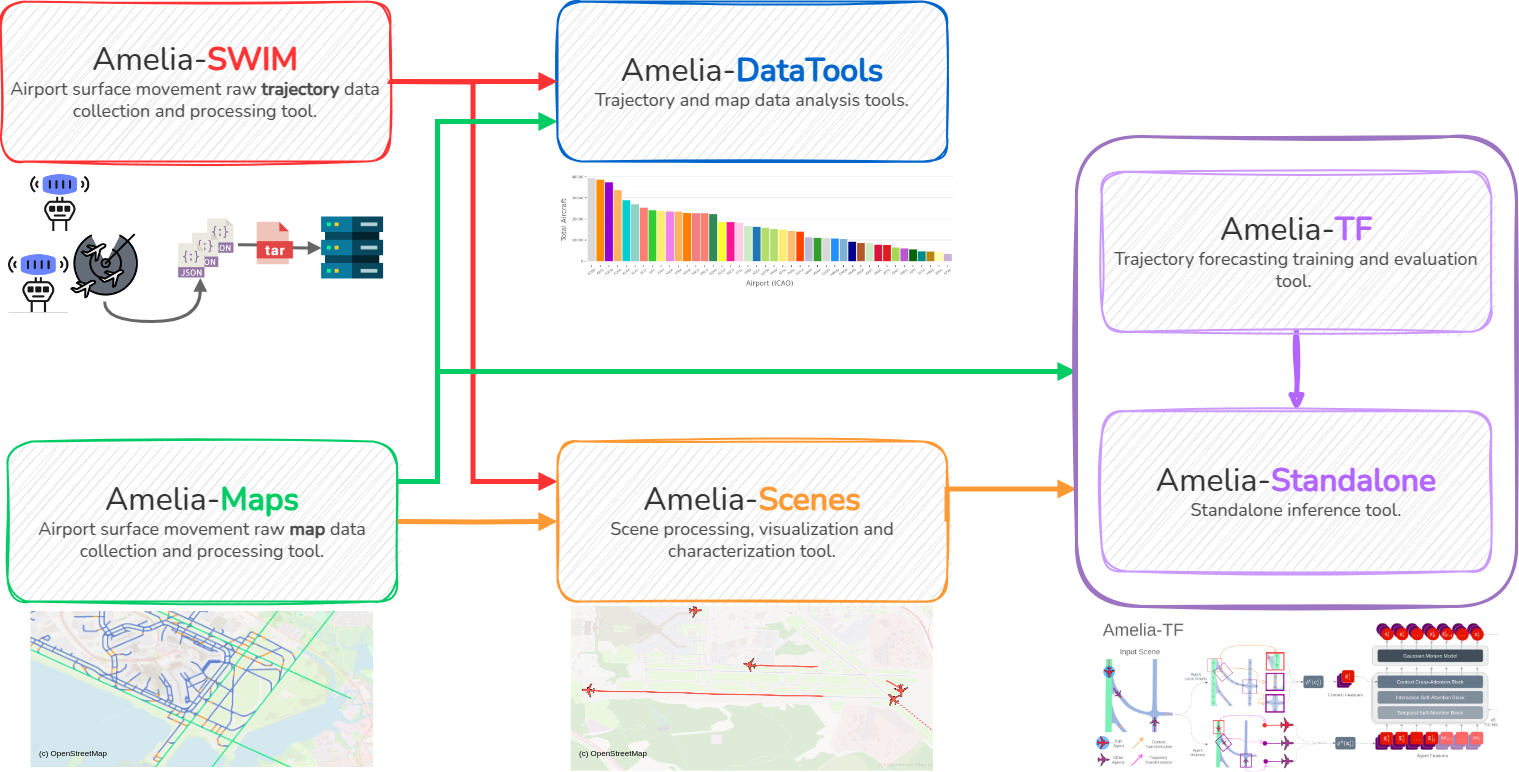}
    \caption{Overview of the \framework~Framework.}
    \label{fig:framework}
\end{figure*}

The \framework~framework shown in \Cref{fig:framework} consists of a toolkit utilized for (1) collecting, visualizing, analyzing, and processing airport trajectory and map data, and (2) developing, training, and evaluating motion forecasting algorithms.  We open-source all of these tools as a set of six GitHub repositories to facilitate data-driven research in the aviation domain at \texttt{\href{https://github.com/orgs/AmeliaCMU/repositories}{github.com/orgs/AmeliaCMU}}.

Below, we provide a brief description of each of these repositories. Further sections in this supplementary provide more details about (1) data collection and processing (\Cref{asec:dataset}); (2) in-depth analysis of our subsets \ameliasubset~(\Cref{asec:subset}) and \ameliabench~(\Cref{asec:benchmark_extra}); and (3) model design and additional experiments~(\Cref{asec:model}). 

\subsection{Amelia-SWIM} \label[appendix]{assec:swim_repo}

This tool provides scripts to \textbf{download} the raw trajectory data described in \Cref{ssec:trajectory_data}, and to \textbf{process} the data into tabular form with the attributes described in \Cref{tab:processed_data}. It also provides instructions on how to modify airport geo-fences. This enables customs regions to be processed in the vicinity of the airport.

The tool and usage instructions are available at:
\texttt{\href{https://github.com/AmeliaCMU/AmeliaSWIM}{github.com/AmeliaCMU/AmeliaSWIM}}. 

\subsection{Amelia-Maps} \label[appendix]{assec:maps_repo}

This tool provides scripts to \textbf{process} the raw OpenStreetMap (OSM) \cite{map2014open} into the vectorized representation utilized in our work, and to visualize the process step-by-step. It also provides a script to obtain the OSM background maps utilized for visualization generation in this paper. 

The tool and usage instructions are available at:
\texttt{\href{https://github.com/AmeliaCMU/AmeliaMaps}{github.com/AmeliaCMU/AmeliaMaps}}. 

\subsection{Amelia-DataTools} \label[appendix]{assec:data_repo}

We offer scripts for analyzing and visualizing map and trajectory data. These scripts include tools for assessing agent counts at each airport (\Cref{fig:teaser}, \Cref{fig:dataset_analysis}), profiling motion (\Cref{fig:statistic_summary}, visualizing trajectory data over a corresponding map (\Cref{fig:dataset_overview}), dataset summaries (\Cref{tab:trajectory_analysis}, \Cref{tab:trajectory_analysis_42}, \Cref{tab:map_analysis}), and more. 

The tool and usage instructions are available at:
\texttt{\href{https://github.com/AmeliaCMU/AmeliaDataTools}{github.com/AmeliaCMU/AmeliaDataTools}}.

\subsection{Amelia-Scenes} \label[appendix]{assec:scenes_repo}

This tool is used for extracting scenes from the tabular data. Here scene is defined as a window of time in which interactions are analyzed. Our scene extraction is easily configurable to obtain different types of scenes in terms of the number of agents of interest, scene length, and granularity. It also provides filtering options to remove scenes with vehicles or unknown agents, if desired. We also provide scene characterization tools to analyze scene complexity \wrt~individual agent kinematic profiles, as well as the level of agent-to-agent interactivity and crowdedness. Finally, we provide scripts with various strategies to split the data into \texttt{train/val/test} sets. 

Scene examples are shown in \Cref{fig:teaser}, \Cref{fig:ego_results} and \Cref{fig:ego_selection}. 

The tool and usage instructions are available at:
\texttt{\href{https://github.com/AmeliaCMU/AmeliaScenes}{github.com/AmeliaCMU/AmeliaScenes}}. 

\subsection{Amelia-TF} \label[appendix]{assec:forecasting_repo}

We provide a framework for training and evaluating trajectory forecasting models that uses Hydra + Pytorch Lightning. This framework enables easy integration and implementation of existing and new models. We also release a standalone inference tool for visualizing model predictions.

The model architecture, configuration files, training/evaluation tools, and usage instructions are available at:
\texttt{\href{https://github.com/AmeliaCMU/AmeliaTF}{github.com/AmeliaCMU/AmeliaTF}}. 

\subsection{Amelia-Inference} \label[appendix]{assec:inference_repo}

This tool is intended to be a standalone version of Amelia-{\color{Orchid}TF}, used only for inference purposes without the training/evaluation and prototyping overhead code from the latter repo. 

The tools and usage instructions are available at:
\texttt{\href{https://github.com/AmeliaCMU/AmeliaInference}{github.com/AmeliaCMU/AmeliaInference}}. 

%% file: appendix/2_dataset.tex
\section{\ameliadataset~Supplementary} \label[appendix]{asec:dataset}

Here, we describe in more detail the collection and processing steps we followed to create the \ameliadataset~dataset. 

\subsection{Collecting the Raw Trajectory Data through SWIM} \label[appendix]{assec:swim}

\ameliadataset~leverages the FAA's System Wide Information Management (SWIM) Program \cite{robb2014system}, a repository of aviation data spanning the National Airspace System (NAS). Specifically, we utilize the SWIM Terminal Data Distribution System (STDDS), which aggregates terminal data from various sources like the Airport Surface Detection Equipment -- Model X (ADSE-X) and the Airport Surface Surveillance Capability (ASSC) into easily accessible information. Position reports for zones covered by STDDS are made available as Surface Movement Events, streamed as a subscription via the NAS Enterprise Messaging Service (NEMS). These events are structured in an XML-like format, as shown in the example in \Cref{lst:raw_SMES}.

STDDS data is only available for airports that are covered by a Terminal Radar Approach Control (TRACON) facility. As a result, the raw dataset is a collection of  Surface Movement Events covering TRACONs within the NAS since December 1st, 2022 (Unix Time Stamp: \verb+1669944602+). The \ameliadataset~dataset continues to stream and archive these position reports on an ongoing basis.

Our recording scripts use two sets of 4 concurrent Advanced Message Queuing Protocol \cite{vinoski2006advanced} connections for redundancy. Each connection records data for one hour and then closes the connection to offload the data to storage as a compressed newline JSON. The format of the files follows the structure: \verb+ALL_<#connection>_<UnixTimeStamp>.njson.gz+. Every hour, the first set starts recording on the hour, and the next one starts 30 minutes later. Thus, at any given point, eight connections are active. Each connection deals with a manageable chunk of data to reduce onboard RAM usage and guard against data corruption. The scripts launch afresh automatically every 30 minutes, providing redundancy against server downtime and networking snags. 

\newpage
\begin{lstlisting}[language=json, caption={Sample of Surface Movement Event messages which are aggregated to form the raw dataset.},captionpos=b, label={lst:raw_SMES}]
{"properties": {"blockAtFAA": "false", "msgType": "ML", "sendTo": "all", "solace_routing_dest_name": "SMES/all/false/ML/KBOS/A90", "globalID": "1453771078", "msgSeqID": "157693832", "tracon": "A90", "timeline": "1736145001439 1736145001519 1736145001519", "version": "4.0", "airport": "KBOS", "timestamp": "2025-01-06T06:30:01.439Z"}, "body": "<?xml version=\"1.0\" encoding=\"UTF-8\" standalone=\"yes\"?><ns2:asdexMsg xmlns:ns2=\"urn:us:gov:dot:faa:atm:terminal:entities:v4-0:smes:surfacemovementevent\"><airport>KBOS</airport><mlatReport full=\"true\"><report><basicReport><time>2025-01-06T06:30:01.250Z</time><track>3328</track><position><x>248</x><y>79</y><lat>42.36647</lat><lon>-71.01542</lon></position><velocity><x>0.0</x><y>0.0</y></velocity></basicReport><acAddress>A70F11</acAddress><height>-256.25</height></report><descriptor><crt>0</crt><rab>0</rab><spi>0</spi><gbs>1</gbs><tot>undetermined</tot><type>1</type></descriptor><status><cnf>confirmed</cnf><dou>1</dou></status><extent><u>0</u><x>0</x><gm>0</gm><df>17</df><xxcovar>2.200000047683716</xxcovar><yycovar>3.9000000953674316</yycovar><zzcovar>100.0</zzcovar><xycovar>-5.0</xycovar></extent><cachedData><mode3ACode>2000</mode3ACode><acAddress>A70F11</acAddress></cachedData></mlatReport><mlatReport full=\"false\"><report><basicReport><time>2025-01-06T06:30:01.289Z</time><track>3423</track><position><x>-277</x><y>-452</y><lat>42.3617</lat><lon>-71.02181</lon></position></basicReport><mode3ACode><code>7160</code><g>0</g></mode3ACode></report><descriptor><crt>0</crt></descriptor><extent><df>5</df><xxcovar>2.799999952316284</xxcovar></extent><enhancedData><eramGufi>KT08053200</eramGufi></enhancedData></mlatReport></ns2:asdexMsg>"}
\end{lstlisting}

\subsection{Geofence Creation} \label[appendix]{assec:fences}

\begin{figure*}[ht!]
    \centering
    \includegraphics[width=0.98\textwidth]{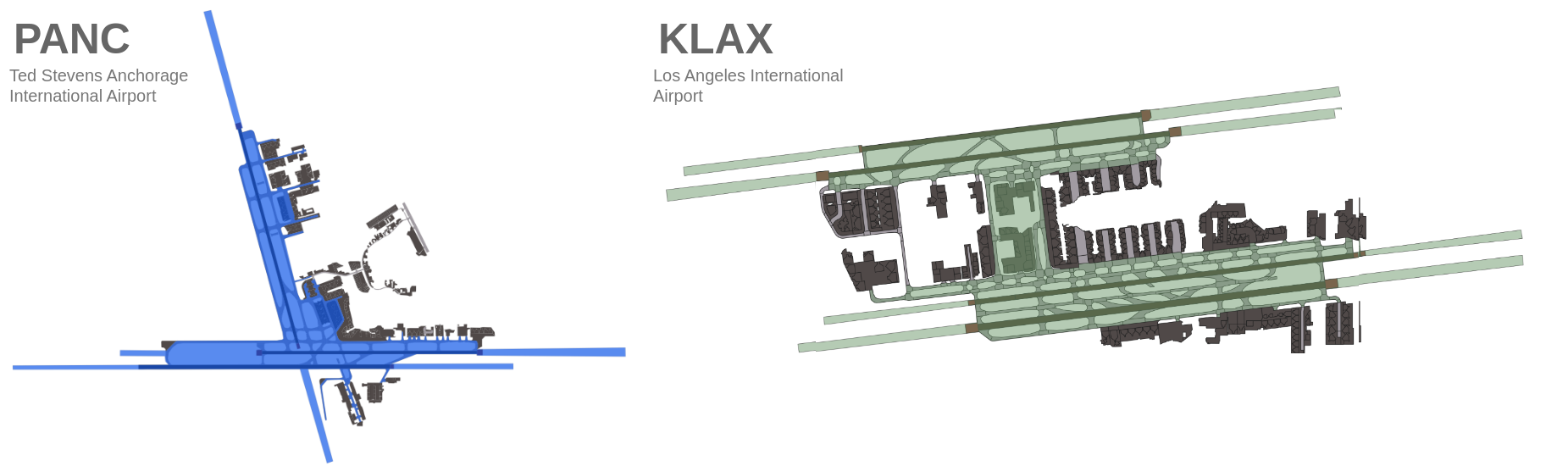}
    \caption{Examples of geofences used for determining a region of interest to collect position reports within the airports' movement areas.}
    \label{fig:fences}
\end{figure*}

Before pre-processing the trajectory data into clean, tabular data suitable for motion forecasting, we need to define a region of interest representing the \textbf{surface movement areas} described in \Cref{sec:dataset}. To do so, we define a 2D geographical fence around the airport of interest and then filter the position reports that fall inside it and within 2000ft above ground level. A few examples of GeoFences resulting from this process are shown in \Cref{fig:fences}.

\subsection{Automated Graph Generation} \label[appendix]{assec:graph_pipeline}

\begin{figure*}[th!]
    \centering
    \includegraphics[width=0.98\textwidth]{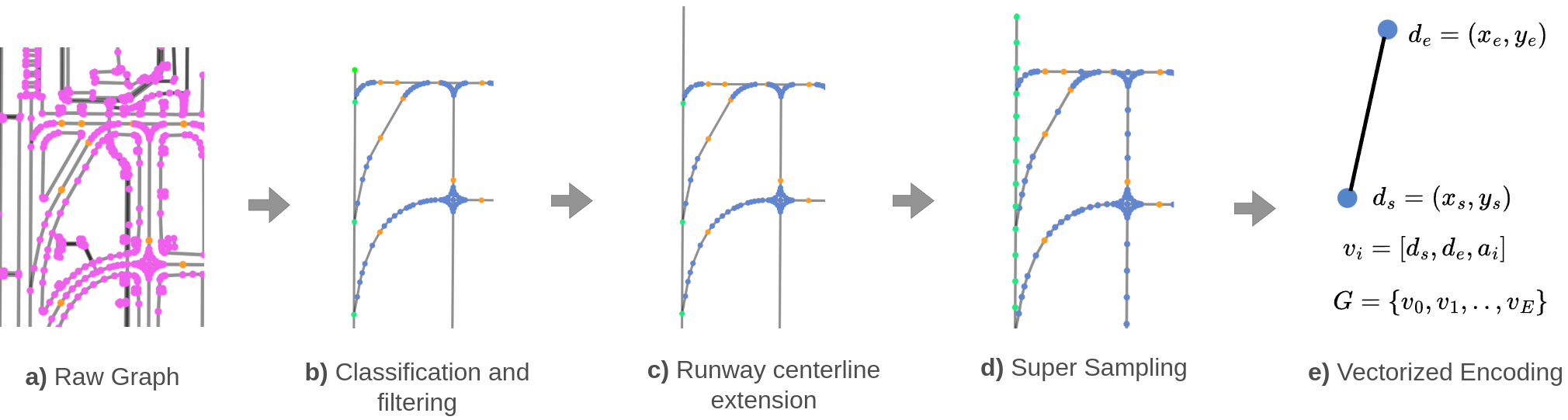}
    \caption{Overview of our automated map processing pipeline, which converts raw maps from OpenStreetMap into compact, easy-to-use semantic airport graphs. }
    \label{fig:graph_processing}
\end{figure*}

To obtain a more compact representation of the map, we apply a series of automated steps to only maintain elements of interest. We begin by filtering the graph such that only edges and nodes pertaining to centerlines of movement areas are maintained. These are then classified into the following semantic classes: runway, taxiway, and hold-lines, which indicate the exit boundary of runway-protected areas (\Cref{fig:graph_processing}.b). We then extend runway endpoints by a nautical mile to encompass landing and take-off procedures (\Cref{fig:graph_processing}.c). Lastly, we super-sample the runway centerlines to obtain a finer runway representation (\Cref{fig:graph_processing}.d).

Finally, following \cite{gao2020vectornet}, we encode the graph in a vectorized manner (\Cref{fig:graph_processing}.e), where an edge $i$ is described as a vector, $v_i = [d_i^s, d_i^e, a_i]$, with $d_i^s\in \R^{1\times4}$ and $d_i^e \in \R^{1\times4}$ being the start and end nodes containing the geographical and relative coordinates, and $a_i \in \N^{1\times3}$ a one-hot encoded vector indicating one of semantic classes. 

%% file: appendix/3_dataset_analysis.tex
\section{\ameliasubset~Supplementary} \label[appendix]{asec:subset}

Here, we provide additional analyses for the processed \ameliasubset~subset.

\subsection{Dataset Analysis} \label[appendix]{assec:subset_traj_analysis}

In \Cref{fig:amelia42_timespan} we show the selected per-airport timeframes for the \ameliasubset~subset. On the horizontal axis, we show the time span of the full \ameliadataset~dataset as of May 22nd, 2025. To obtain the \ameliasubset~subset, and reduce selection biases, we randomly select a starting point (left-most point in each airport's interval) across the timeline for each airport, and process the data for 15 subsequent days from the starting point (right-most point in the interval). Therefore, we show each airport's sample interval in the figure and the airport's ICAO code on the left. The legend includes the specific timeframes as a range \texttt{YYYY-MM-DD - YYYY-MM-DD} reported in the GMT-4 timezone. 

\begin{figure}[!ht]
    \centering
    \includegraphics[width=0.98\linewidth]{figures/amelia42_timespan_compressed.png}
    \caption{Selected per-airport timeframes for the \ameliasubset~subset. }
    \label{fig:amelia42_timespan}
\end{figure}

\clearpage

\Cref{tab:trajectory_analysis_42} provides a quantitative summary of each airport's traffic information for the \ameliasubset~subset. It includes the \textit{total} and \textit{per-category} number of unique agents and unique data points. Like \Cref{fig:dataset_analysis}, we include for reference the timespan for the sample we computed the counts, following the format \texttt{YYYY-MM-DD HH:MM - YYYY-MM-DD HH:MM} also reported in GMT-4 time. For each column, we color the cells to indicate relative scale \wrt maximum value in the column. 

\input{tables/traj_analysis_42}

\clearpage

\begin{figure}[!ht]
    \centering
    \includegraphics[width=1.0\linewidth]{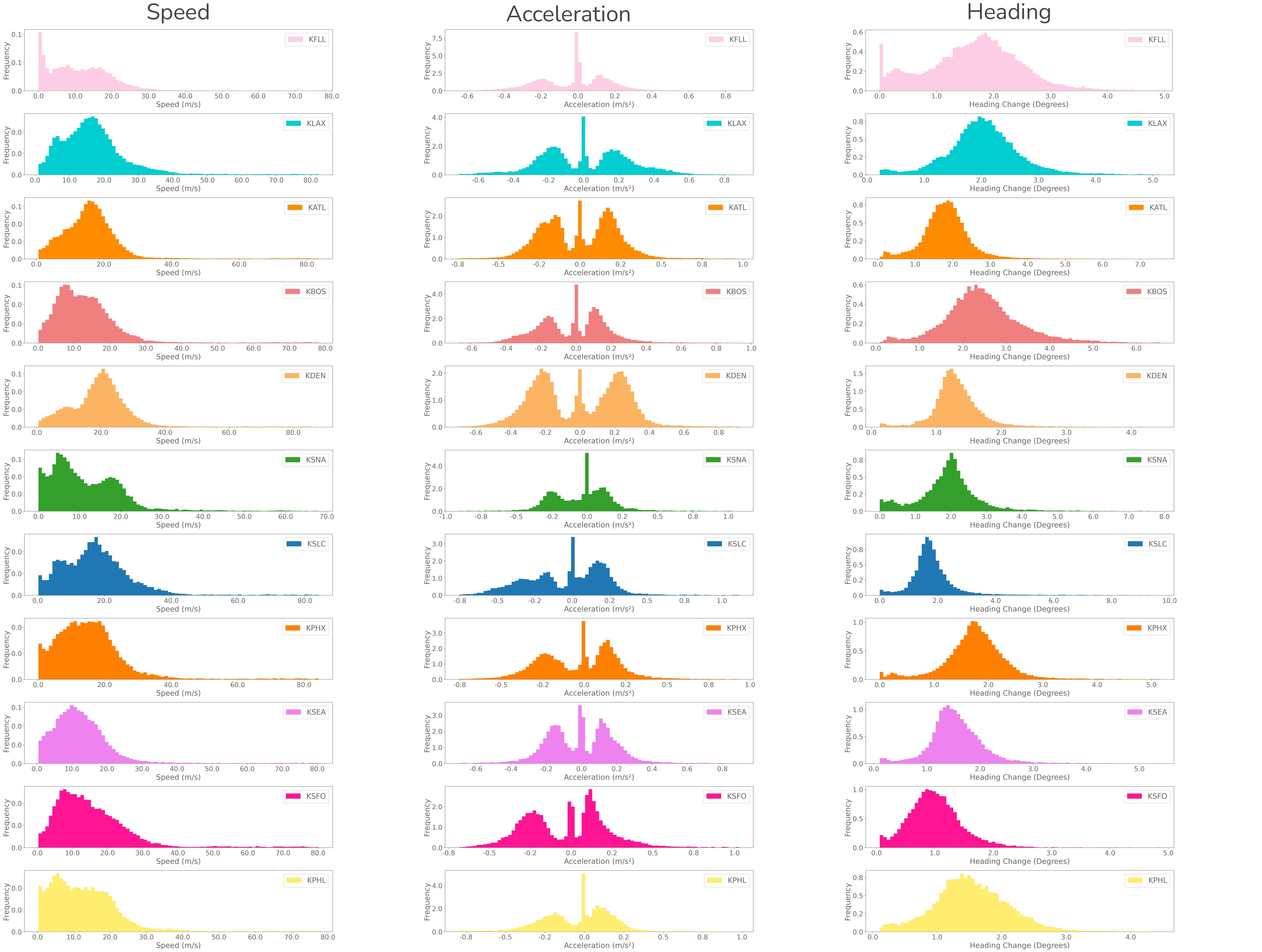}
    \caption{\textbf{Per-airport motion profiles.} We show the per-timestep distribution over the aircraft's speed, acceleration and heading. From top to bottom, we show the motion profiles for: \icao{KFLL}, \icao{KLAX}, \icao{KATL}, \icao{KBOS}, \icao{KDEN}, \icao{KSNA}, \icao{KSLC}, \icao{KPHX}, \icao{KSEA}, \icao{KSFO}, \icao{KPHL}.}
    \label{fig:amelia42_motion_prof1}
\end{figure}

\begin{figure}[H]
    \centering
    \includegraphics[width=1\linewidth]{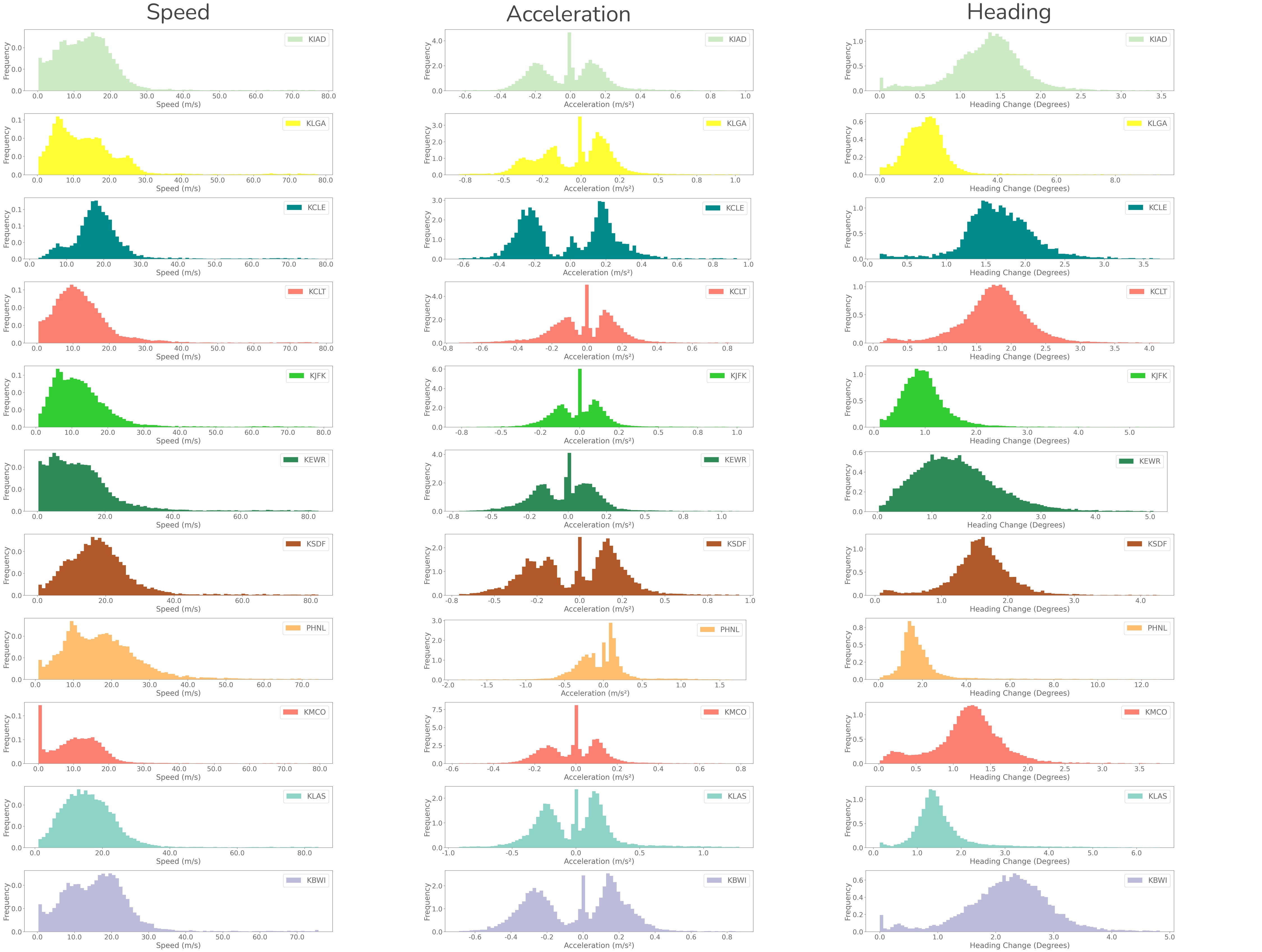}
    \caption{\textbf{Per-airport motion profiles.} We show the per-timestep distribution over the aircraft's speed, acceleration and heading. From top to bottom, we show the motion profiles for: \icao{KIAD}, \icao{KLGA}, \icao{KCLE}, \icao{KCLT}, \icao{KJFK}, \icao{KEWR}, \icao{KSDF}, \icao{PHNL}, \icao{KMCO}, \icao{KLAS}, \icao{KBWI}. }
    \label{fig:amelia42_motion_prof2}
\end{figure}

\begin{figure}[H]
    \centering
    \includegraphics[width=1.0\linewidth]{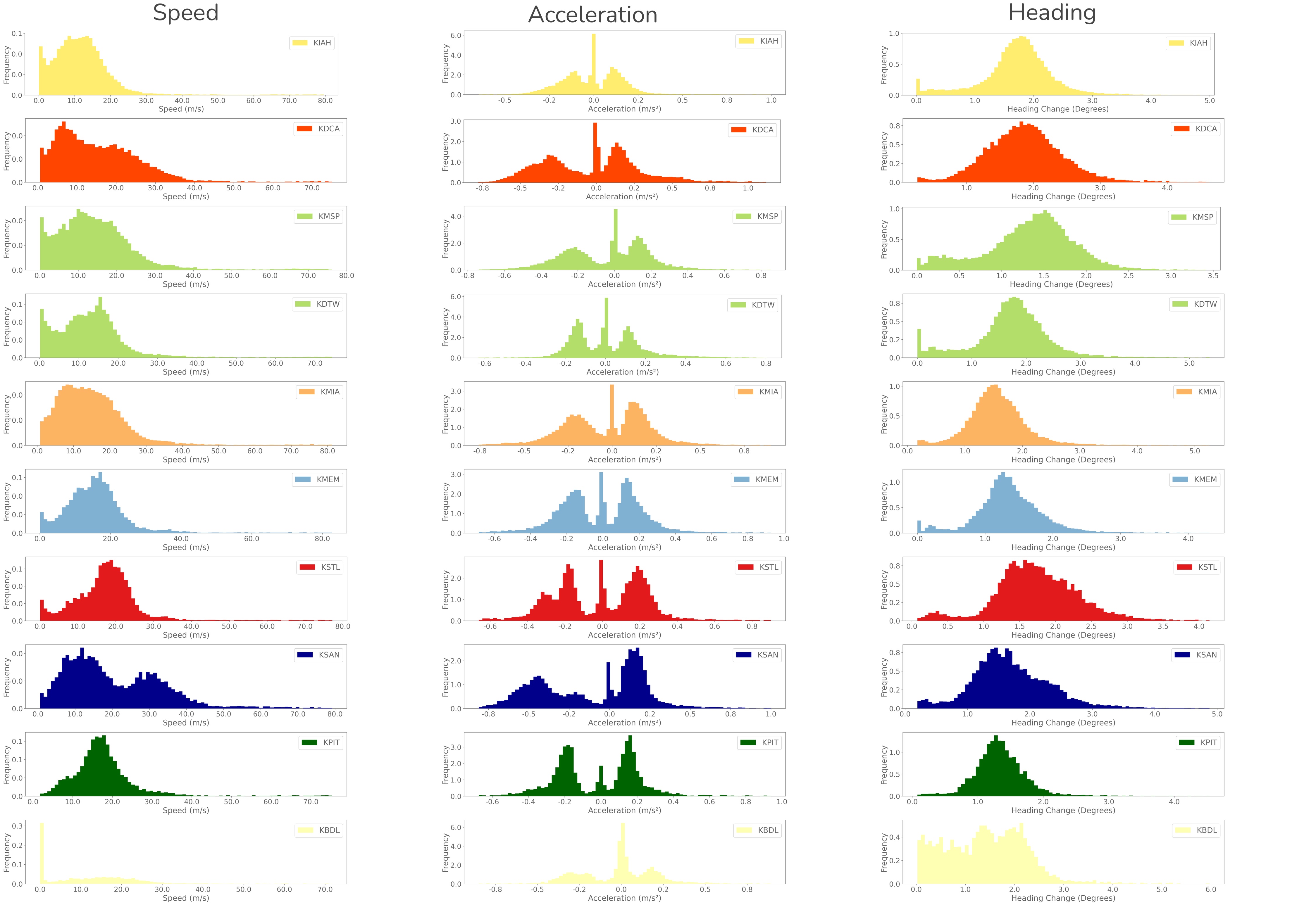}
     \caption{\textbf{Per-airport motion profiles.} We show the per-timestep distribution over the aircraft's speed, acceleration and heading. From top to bottom, we show the motion profiles for: \icao{KIAH}, \icao{KDCA}, \icao{KMSP}, \icao{KDTW}, \icao{KMIA}, \icao{KMEM}, \icao{KSTL}, \icao{KSAN}, \icao{KPIT}, \icao{KBDL}.}
    \label{fig:amelia42_motion_prof3}
\end{figure}

\begin{figure}[H]
    \centering
    \includegraphics[width=1.0\linewidth]{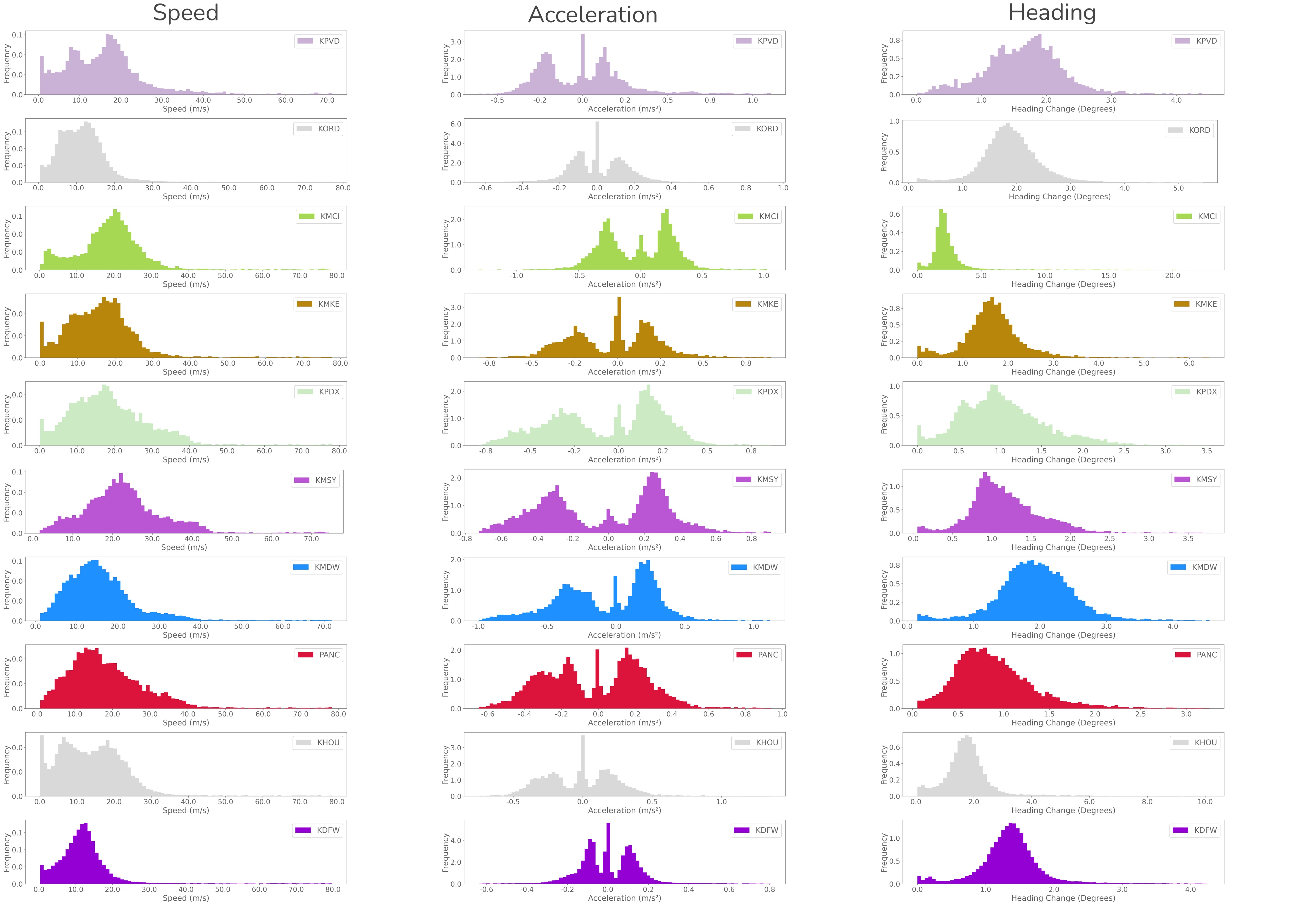}
    \caption{\textbf{Per-airport motion profiles.} We show the per-timestep distribution over the aircraft's speed, acceleration and heading. From top to bottom, we show the motion profiles for: \icao{KPVD}, \icao{KORD}, \icao{KMCI}, \icao{KMKE}, \icao{KPDX}, \icao{KMSY}, \icao{KMDW}, \icao{PANC}, \icao{KHOU}, \icao{KDFW}.}
    \label{fig:amelia42_motion_prof4}
\end{figure}

In \Cref{fig:amelia42_motion_prof1} through \ref{fig:amelia42_motion_prof4}, we present the per-timestep motion profiles of aircraft across all airports in the subset. From left to right, the plots display average speed, average acceleration, and heading change, with each row representing a different airport. These profiles reveal ample variability in speed and heading values, capturing the range of aircraft behaviors such as taxiing, take-off, landing, and vacating, as well as behavioral diversity resulting from topological variability.

The acceleration profiles feature a thin, elongated bin at 0m/s\textsuperscript{2}, which corresponds, for instance, to stationary aircraft awaiting clearance or gate assignment, or agents moving at constant speed. Meanwhile, the higher acceleration and deceleration values align with aircraft during take-off and landing. Despite differences between airports, the trends in speed, acceleration, and heading change remain consistent, suggesting that \ameliasubset~successfully captures representative and realistic ground operation behaviors.

\vspace{0.5cm}

\clearpage

\clearpage

\begin{figure}[H]
    \centering
    \includegraphics[width=0.75\linewidth]{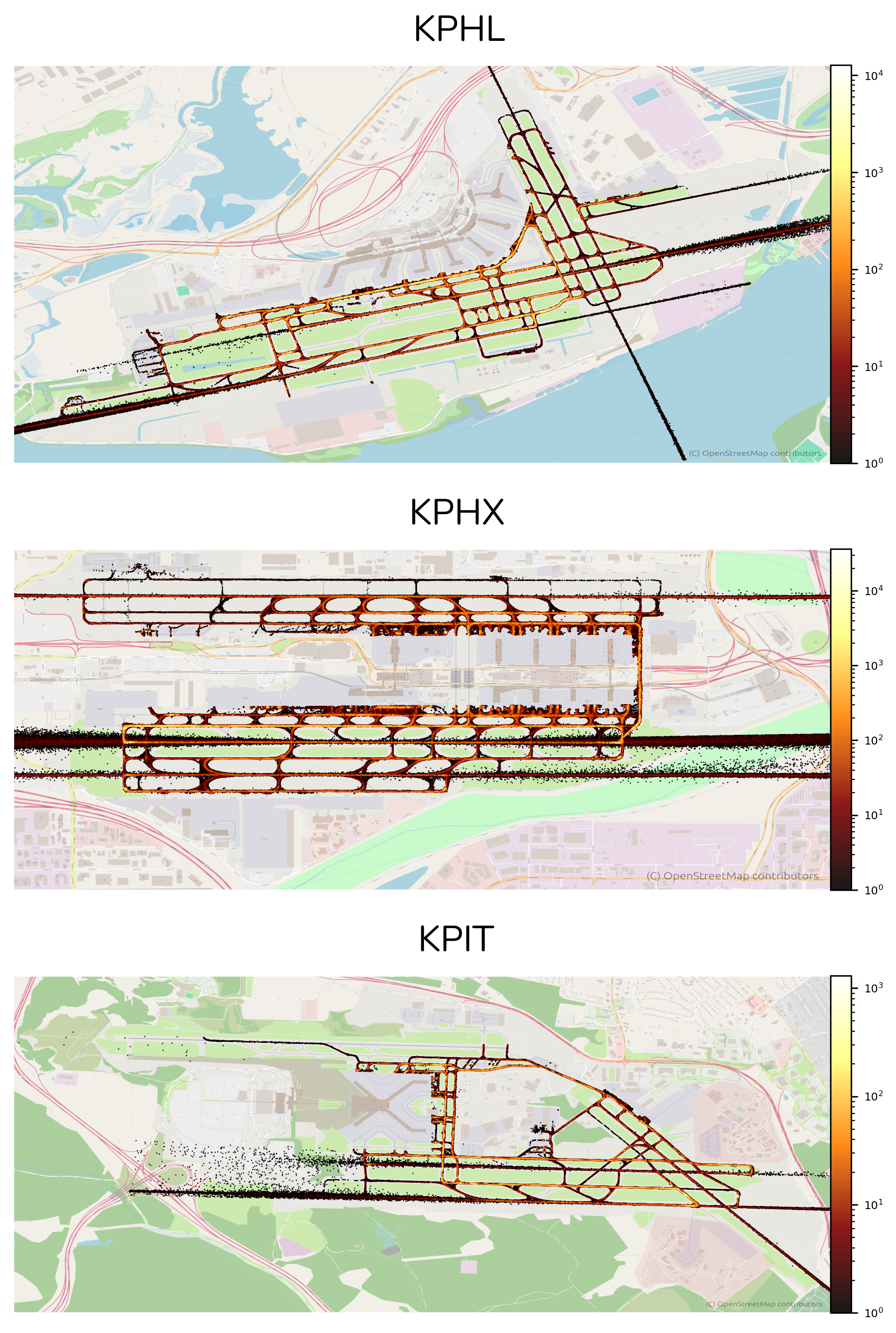}
    \caption{Heatmap showing activity frequency for regions for airports \icao{KPHL}, \icao{KPHX} and \icao{KPIT}}
    \label{fig:amelia42_heatmap1}
\end{figure}

\clearpage

\begin{figure}[H]
    \centering
    \includegraphics[width=0.75\linewidth]{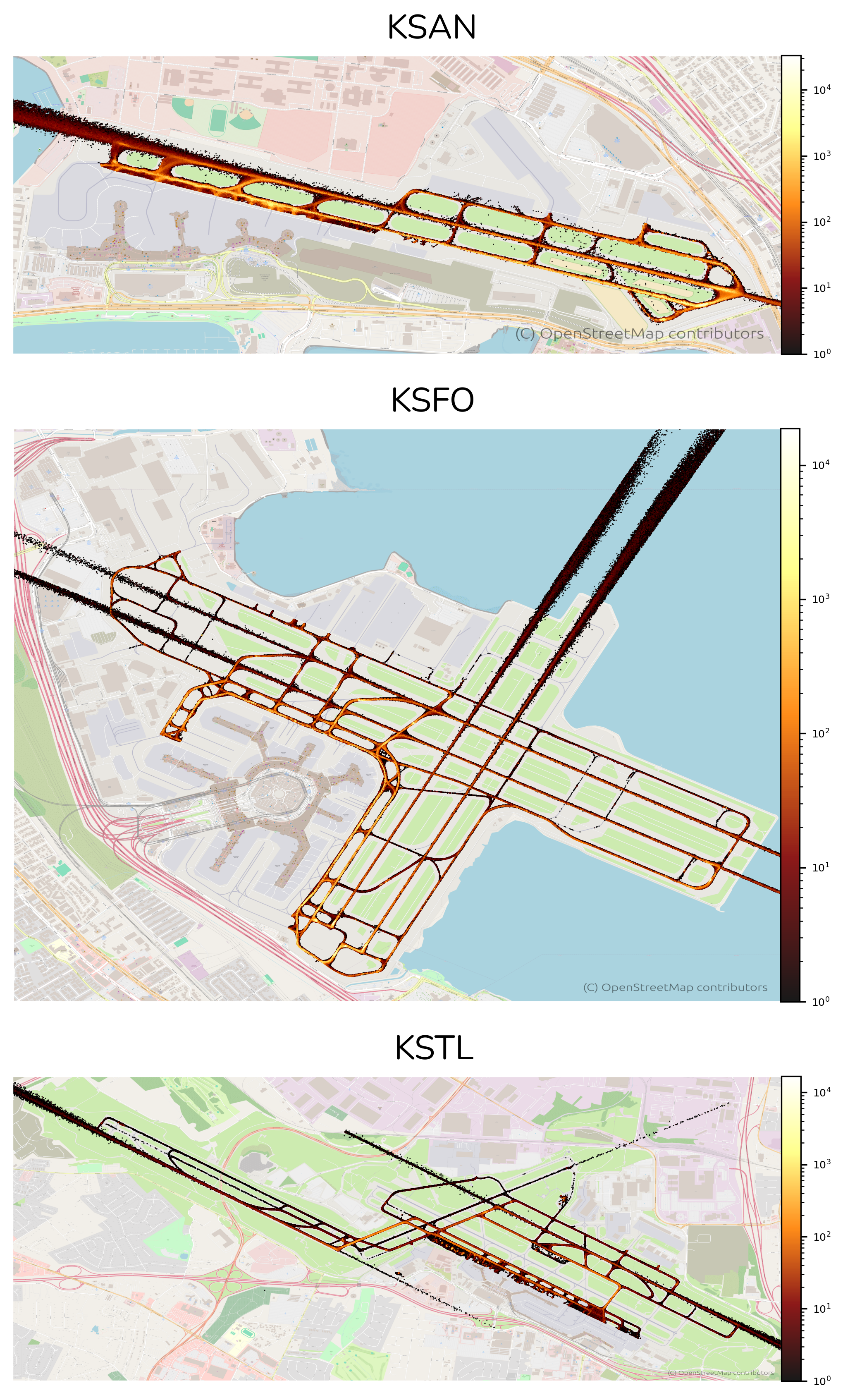}
    \caption{Heatmap showing activity frequency for regions for airports \icao{KSAN}, \icao{KSFO} and \icao{KSTL}}
    \label{fig:amelia42_heatmap2}
\end{figure}

%% file: tables/traj_analysis_42.tex
\begin{table*}[h!]
\caption{Quantitative summaries for the \ameliasubset~subset. }
\label{tab:trajectory_analysis_42}
\resizebox{0.98\textwidth}{!}{
\begin{tabular}{llrrrrrrrr}
\toprule
\multicolumn{2}{c}{} & 
\multicolumn{2}{c}{\textbf{Total}} & 
\multicolumn{2}{c}{\textbf{\aircraft}} & 
\multicolumn{2}{c}{\textbf{\vehicle}} & 
\multicolumn{2}{c}{\textbf{\unknown}} \\[0.1em]
\cmidrule(lr){3-4}
\cmidrule(lr){5-6}
\cmidrule(lr){7-8}
\cmidrule(lr){9-10}
\multicolumn{1}{c}{\textbf{Airport}} & 
\multicolumn{1}{c}{\textbf{Timespan}} &
\textbf{Num. Agents} & \textbf{Num. Data Points} & 
\textbf{Num. Agents} & \textbf{Num. Data Points} & 
\textbf{Num. Agents} & \textbf{Num. Data Points} & 
\textbf{Num. Agents} & \textbf{Num. Data Points} \\
\midrule
\icao{KATL} & 2024-06-21 19:04 - 2024-07-06 18:04 & \cellcolor{Gray!45}410.11K & \cellcolor{Gray!42}37.89M & \cellcolor{Red!97}38.67K & \cellcolor{Red!68}16.02M & \cellcolor{RoyalBlue!31}3.77K & \cellcolor{RoyalBlue!17}2.06M & \cellcolor{LimeGreen!42}367.61K & \cellcolor{LimeGreen!29}19.78M \\
\icao{KBDL} & 2023-03-31 20:51 - 2023-04-18 00:51 & \cellcolor{Gray!7}68.60K & \cellcolor{Gray!6}6.02M & \cellcolor{Red!11}4.38K & \cellcolor{Red!5}1.20M & \cellcolor{RoyalBlue!6}801 & \cellcolor{RoyalBlue!2}328.68K & \cellcolor{LimeGreen!7}63.42K & \cellcolor{LimeGreen!6}4.49M \\
\icao{KBOS} & 2024-09-26 12:11 - 2024-10-11 14:11 & \cellcolor{Gray!9}87.44K & \cellcolor{Gray!23}21.11M & \cellcolor{Red!57}22.69K & \cellcolor{Red!45}10.61M & \cellcolor{RoyalBlue!24}2.99K & \cellcolor{RoyalBlue!22}2.63M & \cellcolor{LimeGreen!7}61.76K & \cellcolor{LimeGreen!11}7.86M \\
\icao{KBWI} & 2023-12-08 02:04 - 2023-12-23 01:04 & \cellcolor{Gray!4}45.36K & \cellcolor{Gray!9}8.08M & \cellcolor{Red!28}11.22K & \cellcolor{Red!17}4.16M & \cellcolor{RoyalBlue!0}118 & \cellcolor{RoyalBlue!0}50.51K & \cellcolor{LimeGreen!3}34.02K & \cellcolor{LimeGreen!5}3.87M \\
\icao{KCLE} & 2023-05-10 00:59 - 2023-05-30 12:59 & \cellcolor{Gray!3}35.88K & \cellcolor{Gray!4}4.39M & \cellcolor{Red!11}4.52K & \cellcolor{Red!6}1.49M & \cellcolor{RoyalBlue!12}1.52K & \cellcolor{RoyalBlue!12}1.41M & \cellcolor{LimeGreen!3}29.85K & \cellcolor{LimeGreen!2}1.48M \\
\icao{KCLT} & 2023-02-18 21:58 - 2023-03-05 20:58 & \cellcolor{Gray!15}144.91K & \cellcolor{Gray!32}28.77M & \cellcolor{Red!64}25.32K & \cellcolor{Red!55}13.01M & \cellcolor{RoyalBlue!36}4.41K & \cellcolor{RoyalBlue!24}2.86M & \cellcolor{LimeGreen!13}115.17K & \cellcolor{LimeGreen!18}12.90M \\
\icao{KDCA} & 2023-06-12 01:33 - 2023-06-27 00:33 & \cellcolor{Gray!4}37.15K & \cellcolor{Gray!11}10.04M & \cellcolor{Red!35}13.97K & \cellcolor{Red!20}4.68M & \cellcolor{RoyalBlue!7}930 & \cellcolor{RoyalBlue!8}952.03K & \cellcolor{LimeGreen!2}22.24K & \cellcolor{LimeGreen!6}4.40M \\
\icao{KDEN} & 2025-02-25 11:48 - 2025-03-13 05:48 & \cellcolor{Gray!25}229.60K & \cellcolor{Gray!22}20.24M & \cellcolor{Red!85}33.63K & \cellcolor{Red!51}12.05M & \cellcolor{RoyalBlue!26}3.17K & \cellcolor{RoyalBlue!15}1.82M & \cellcolor{LimeGreen!22}192.80K & \cellcolor{LimeGreen!9}6.38M \\
\icao{KDFW} & 2024-02-12 09:33 - 2024-02-27 18:33 & \cellcolor{Gray!35}320.53K & \cellcolor{Gray!100}88.77M & \cellcolor{Red!94}37.45K & \cellcolor{Red!100}23.40M & \cellcolor{RoyalBlue!0}5 & \cellcolor{RoyalBlue!0}2.11K & \cellcolor{LimeGreen!32}283.07K & \cellcolor{LimeGreen!96}65.36M \\
\icao{KDTW} & 2024-02-24 10:25 - 2024-03-12 11:25 & \cellcolor{Gray!23}212.34K & \cellcolor{Gray!32}28.69M & \cellcolor{Red!39}15.78K & \cellcolor{Red!32}7.57M & \cellcolor{RoyalBlue!36}4.43K & \cellcolor{RoyalBlue!21}2.55M & \cellcolor{LimeGreen!22}192.13K & \cellcolor{LimeGreen!27}18.56M \\
\icao{KEWR} & 2024-06-14 01:35 - 2024-06-29 03:35 & \cellcolor{Gray!11}101.42K & \cellcolor{Gray!23}21.16M & \cellcolor{Red!56}22.28K & \cellcolor{Red!57}13.37M & \cellcolor{RoyalBlue!18}2.25K & \cellcolor{RoyalBlue!12}1.43M & \cellcolor{LimeGreen!8}76.88K & \cellcolor{LimeGreen!9}6.36M \\
\icao{KFLL} & 2025-01-14 20:48 - 2025-01-29 19:48 & \cellcolor{Gray!25}235.95K & \cellcolor{Gray!29}26.42M & \cellcolor{Red!46}18.20K & \cellcolor{Red!43}10.19M & \cellcolor{RoyalBlue!23}2.83K & \cellcolor{RoyalBlue!16}1.92M & \cellcolor{LimeGreen!25}214.91K & \cellcolor{LimeGreen!21}14.30M \\
\icao{KHOU} & 2023-12-06 15:20 - 2023-12-21 14:20 & \cellcolor{Gray!7}64.90K & \cellcolor{Gray!11}9.80M & \cellcolor{Red!27}10.89K & \cellcolor{Red!13}3.23M & \cellcolor{RoyalBlue!3}461 & \cellcolor{RoyalBlue!2}293.23K & \cellcolor{LimeGreen!6}53.55K & \cellcolor{LimeGreen!9}6.29M \\
\icao{KIAD} & 2025-02-20 11:06 - 2025-03-08 03:06 & \cellcolor{Gray!35}327.11K & \cellcolor{Gray!40}36.01M & \cellcolor{Red!42}16.63K & \cellcolor{Red!32}7.72M & \cellcolor{RoyalBlue!34}4.16K & \cellcolor{RoyalBlue!12}1.45M & \cellcolor{LimeGreen!35}306.32K & \cellcolor{LimeGreen!39}26.83M \\
\icao{KIAH} & 2024-11-28 00:45 - 2024-12-12 23:45 & \cellcolor{Gray!39}363.85K & \cellcolor{Gray!68}61.22M & \cellcolor{Red!60}23.87K & \cellcolor{Red!57}13.50M & \cellcolor{RoyalBlue!66}8.05K & \cellcolor{RoyalBlue!33}3.87M & \cellcolor{LimeGreen!38}331.90K & \cellcolor{LimeGreen!64}43.79M \\
\icao{KJFK} & 2024-12-12 04:13 - 2024-12-27 03:13 & \cellcolor{Gray!67}615.27K & \cellcolor{Gray!39}35.24M & \cellcolor{Red!60}24.11K & \cellcolor{Red!60}14.19M & \cellcolor{RoyalBlue!0}44 & \cellcolor{RoyalBlue!0}20.53K & \cellcolor{LimeGreen!68}591.11K & \cellcolor{LimeGreen!30}21.03M \\
\icao{KLAS} & 2024-04-30 17:31 - 2024-05-15 16:31 & \cellcolor{Gray!8}76.76K & \cellcolor{Gray!16}15.00M & \cellcolor{Red!68}26.98K & \cellcolor{Red!43}10.16M & \cellcolor{RoyalBlue!17}2.05K & \cellcolor{RoyalBlue!6}809.72K & \cellcolor{LimeGreen!5}47.73K & \cellcolor{LimeGreen!5}4.03M \\
\icao{KLAX} & 2024-09-03 18:42 - 2024-09-19 04:42 & \cellcolor{Gray!70}643.28K & \cellcolor{Gray!55}49.02M & \cellcolor{Red!72}28.81K & \cellcolor{Red!46}10.90M & \cellcolor{RoyalBlue!80}9.72K & \cellcolor{RoyalBlue!20}2.42M & \cellcolor{LimeGreen!70}604.72K & \cellcolor{LimeGreen!52}35.69M \\
\icao{KLGA} & 2024-10-28 08:09 - 2024-11-12 07:09 & \cellcolor{Gray!18}171.71K & \cellcolor{Gray!17}15.90M & \cellcolor{Red!46}18.46K & \cellcolor{Red!30}7.09M & \cellcolor{RoyalBlue!0}65 & \cellcolor{RoyalBlue!0}21.94K & \cellcolor{LimeGreen!17}153.17K & \cellcolor{LimeGreen!12}8.79M \\
\icao{KMCI} & 2024-09-26 15:35 - 2024-10-11 14:35 & \cellcolor{Gray!8}80.81K & \cellcolor{Gray!35}31.22M & \cellcolor{Red!16}6.46K & \cellcolor{Red!7}1.75M & \cellcolor{RoyalBlue!0}44 & \cellcolor{RoyalBlue!0}37.21K & \cellcolor{LimeGreen!8}74.30K & \cellcolor{LimeGreen!43}29.43M \\
\icao{KMCO} & 2023-01-12 08:20 - 2023-01-27 08:20 & \cellcolor{Gray!59}541.20K & \cellcolor{Gray!60}53.29M & \cellcolor{Red!57}22.68K & \cellcolor{Red!58}13.70M & \cellcolor{RoyalBlue!5}644 & \cellcolor{RoyalBlue!2}277.79K & \cellcolor{LimeGreen!60}517.85K & \cellcolor{LimeGreen!57}39.29M \\
\icao{KMDW} & 2023-05-19 21:24 - 2023-06-04 11:24 & \cellcolor{Gray!12}118.06K & \cellcolor{Gray!13}11.82M & \cellcolor{Red!27}10.70K & \cellcolor{Red!12}2.88M & \cellcolor{RoyalBlue!17}2.14K & \cellcolor{RoyalBlue!9}1.15M & \cellcolor{LimeGreen!12}105.22K & \cellcolor{LimeGreen!11}7.78M \\
\icao{KMEM} & 2024-05-28 22:40 - 2024-06-12 23:40 & \cellcolor{Gray!11}105.73K & \cellcolor{Gray!19}16.89M & \cellcolor{Red!26}10.51K & \cellcolor{Red!18}4.44M & \cellcolor{RoyalBlue!6}730 & \cellcolor{RoyalBlue!3}414.54K & \cellcolor{LimeGreen!11}94.49K & \cellcolor{LimeGreen!17}12.04M \\
\icao{KMIA} & 2023-11-22 12:56 - 2023-12-07 11:56 & \cellcolor{Gray!9}85.64K & \cellcolor{Gray!17}15.15M & \cellcolor{Red!59}23.47K & \cellcolor{Red!45}10.71M & \cellcolor{RoyalBlue!22}2.68K & \cellcolor{RoyalBlue!7}894.51K & \cellcolor{LimeGreen!6}59.48K & \cellcolor{LimeGreen!5}3.54M \\
\icao{KMKE} & 2023-06-20 17:40 - 2023-07-05 16:40 & \cellcolor{Gray!28}261.35K & \cellcolor{Gray!39}34.62M & \cellcolor{Red!11}4.46K & \cellcolor{Red!6}1.54M & \cellcolor{RoyalBlue!15}1.86K & \cellcolor{RoyalBlue!12}1.42M & \cellcolor{LimeGreen!29}255.03K & \cellcolor{LimeGreen!46}31.65M \\
\icao{KMSP} & 2023-03-13 02:34 - 2023-04-01 18:34 & \cellcolor{Gray!13}125.92K & \cellcolor{Gray!20}18.23M & \cellcolor{Red!38}15.36K & \cellcolor{Red!28}6.70M & \cellcolor{RoyalBlue!9}1.18K & \cellcolor{RoyalBlue!6}784.28K & \cellcolor{LimeGreen!12}109.38K & \cellcolor{LimeGreen!15}10.75M \\
\icao{KMSY} & 2025-03-26 06:53 - 2025-04-11 01:53 & \cellcolor{Gray!2}20.93K & \cellcolor{Gray!3}3.28M & \cellcolor{Red!15}5.99K & \cellcolor{Red!6}1.47M & \cellcolor{RoyalBlue!0}10 & \cellcolor{RoyalBlue!0}10.39K & \cellcolor{LimeGreen!1}14.94K & \cellcolor{LimeGreen!2}1.80M \\
\icao{KORD} & 2023-10-16 11:52 - 2023-10-31 15:52 & \cellcolor{Gray!100}910.66K & \cellcolor{Gray!81}72.77M & \cellcolor{Red!99}39.53K & \cellcolor{Red!96}22.57M & \cellcolor{RoyalBlue!100}12.02K & \cellcolor{RoyalBlue!99}11.66M & \cellcolor{LimeGreen!99}858.39K & \cellcolor{LimeGreen!55}38.00M \\
\icao{KPDX} & 2024-11-06 04:59 - 2024-11-21 03:59 & \cellcolor{Gray!2}20.47K & \cellcolor{Gray!4}3.72M & \cellcolor{Red!21}8.52K & \cellcolor{Red!11}2.67M & \cellcolor{RoyalBlue!0}119 & \cellcolor{RoyalBlue!0}37.90K & \cellcolor{LimeGreen!1}11.83K & \cellcolor{LimeGreen!1}1.02M \\
\icao{KPHL} & 2023-06-27 11:46 - 2023-07-12 10:46 & \cellcolor{Gray!20}186.58K & \cellcolor{Gray!52}46.32M & \cellcolor{Red!37}14.88K & \cellcolor{Red!36}8.56M & \cellcolor{RoyalBlue!16}1.97K & \cellcolor{RoyalBlue!20}2.36M & \cellcolor{LimeGreen!19}169.73K & \cellcolor{LimeGreen!52}35.41M \\
\icao{KPHX} & 2024-02-01 04:06 - 2024-02-16 03:06 & \cellcolor{Gray!38}347.51K & \cellcolor{Gray!46}41.46M & \cellcolor{Red!57}22.91K & \cellcolor{Red!42}9.88M & \cellcolor{RoyalBlue!5}625 & \cellcolor{RoyalBlue!4}553.10K & \cellcolor{LimeGreen!37}323.98K & \cellcolor{LimeGreen!45}31.02M \\
\icao{KPIT} & 2024-04-11 15:42 - 2024-04-26 14:42 & \cellcolor{Gray!10}99.15K & \cellcolor{Gray!25}22.37M & \cellcolor{Red!13}5.51K & \cellcolor{Red!8}2.04M & \cellcolor{RoyalBlue!9}1.17K & \cellcolor{RoyalBlue!6}816.24K & \cellcolor{LimeGreen!10}92.47K & \cellcolor{LimeGreen!28}19.51M \\
\icao{KPVD} & 2024-07-28 11:34 - 2024-08-12 13:34 & \cellcolor{Gray!4}37.58K & \cellcolor{Gray!3}3.49M & \cellcolor{Red!8}3.33K & \cellcolor{Red!5}1.32M & \cellcolor{RoyalBlue!0}12 & \cellcolor{RoyalBlue!0}6.72K & \cellcolor{LimeGreen!3}34.24K & \cellcolor{LimeGreen!3}2.16M \\
\icao{KSAN} & 2023-01-28 03:49 - 2023-02-12 05:49 & \cellcolor{Gray!1}11.47K & \cellcolor{Gray!4}3.62M & \cellcolor{Red!23}9.19K & \cellcolor{Red!13}3.16M & \cellcolor{RoyalBlue!2}317 & \cellcolor{RoyalBlue!0}112.98K & \cellcolor{LimeGreen!0}1.96K & \cellcolor{LimeGreen!0}351.27K \\
\icao{KSDF} & 2024-08-29 06:01 - 2024-09-13 10:01 & \cellcolor{Gray!26}241.46K & \cellcolor{Gray!30}26.86M & \cellcolor{Red!21}8.57K & \cellcolor{Red!13}3.14M & \cellcolor{RoyalBlue!0}113 & \cellcolor{RoyalBlue!0}106.41K & \cellcolor{LimeGreen!27}232.77K & \cellcolor{LimeGreen!34}23.61M \\
\icao{KSEA} & 2023-06-29 01:01 - 2023-07-14 00:01 & \cellcolor{Gray!18}167.14K & \cellcolor{Gray!19}17.14M & \cellcolor{Red!59}23.57K & \cellcolor{Red!49}11.62M & \cellcolor{RoyalBlue!1}146 & \cellcolor{RoyalBlue!1}149.44K & \cellcolor{LimeGreen!16}143.42K & \cellcolor{LimeGreen!7}5.37M \\
\icao{KSFO} & 2025-02-15 02:09 - 2025-03-02 01:09 & \cellcolor{Gray!9}85.55K & \cellcolor{Gray!14}13.21M & \cellcolor{Red!46}18.42K & \cellcolor{Red!32}7.59M & \cellcolor{RoyalBlue!69}8.40K & \cellcolor{RoyalBlue!26}3.10M & \cellcolor{LimeGreen!6}58.73K & \cellcolor{LimeGreen!3}2.52M \\
\icao{KSLC} & 2024-09-27 21:00 - 2024-10-13 00:00 & \cellcolor{Gray!22}203.57K & \cellcolor{Gray!42}37.34M & \cellcolor{Red!41}16.26K & \cellcolor{Red!26}6.18M & \cellcolor{RoyalBlue!4}524 & \cellcolor{RoyalBlue!1}125.68K & \cellcolor{LimeGreen!21}186.76K & \cellcolor{LimeGreen!45}31.03M \\
\icao{KSNA} & 2024-05-11 21:46 - 2024-05-26 20:46 & \cellcolor{Gray!5}49.37K & \cellcolor{Gray!6}5.66M & \cellcolor{Red!27}11.02K & \cellcolor{Red!19}4.66M & \cellcolor{RoyalBlue!0}45 & \cellcolor{RoyalBlue!0}19.84K & \cellcolor{LimeGreen!4}38.31K & \cellcolor{LimeGreen!1}976.05K \\
\icao{KSTL} & 2023-04-22 04:14 - 2023-05-08 07:14 & \cellcolor{Gray!23}211.96K & \cellcolor{Gray!81}72.38M & \cellcolor{Red!19}7.70K & \cellcolor{Red!12}2.89M & \cellcolor{RoyalBlue!12}1.52K & \cellcolor{RoyalBlue!13}1.54M & \cellcolor{LimeGreen!23}202.73K & \cellcolor{LimeGreen!100}67.95M \\
\icao{PANC} & 2023-12-17 23:41 - 2024-01-01 22:41 & \cellcolor{Gray!6}61.93K & \cellcolor{Gray!9}8.65M & \cellcolor{Red!19}7.80K & \cellcolor{Red!11}2.71M & \cellcolor{RoyalBlue!16}2.03K & \cellcolor{RoyalBlue!12}1.47M & \cellcolor{LimeGreen!6}52.10K & \cellcolor{LimeGreen!6}4.47M \\
\icao{PHNL} & 2023-12-26 16:34 - 2024-01-10 15:34 & \cellcolor{Gray!29}265.90K & \cellcolor{Gray!21}19.40M & \cellcolor{Red!35}14.17K & \cellcolor{Red!22}5.17M & \cellcolor{RoyalBlue!0}96 & \cellcolor{RoyalBlue!0}60.60K & \cellcolor{LimeGreen!29}251.63K & \cellcolor{LimeGreen!20}14.17M \\
\midrule
\rowcolor{Gray!2}
\textbf{Total} &  & \textbf{8.43M} & \textbf{1.10B} & \textbf{708.88K} & \textbf{321.87M} & \textbf{90.14K} & \textbf{54.01M} & \textbf{7.63M} & \textbf{726.04M} \\
\bottomrule
\end{tabular}

}
\end{table*}

%% file: appendix/4_benchmark.tex
\section{\ameliabench~Supplementary} \label[appendix]{asec:benchmark_extra}

\ameliabench~ is meant to provide a diverse set of settings and interactions for benchmarking trajectory forecasting performance. This section provides an analysis of the selected airports to support this objective. 

\subsection{Trajectory Data Analysis.} \label[appendix]{assec:bench_traj_analysis}


\Cref{tab:trajectory_analysis} provides a quantitative summary of each airport's traffic information, including the \textit{total} and \textit{per-category} number of unique agents and unique data points. This validates that the selected airports widely vary in traffic density and overall crowdness. The table includes additional information such as the month from which we sampled the data, and data quality in terms of interpolated points and agent type percentages (\texttt{Interp} and \texttt{Type} in \Cref{tab:processed_data}). For instance, out of the $\sim$300M global data points, only 3.46\% of the data points were interpolated, suggesting minimal need for interpolation. Finally, while we note that the majority of the unique agents ($\sim$83.47\%) are classified as \textit{unknown}, we have no control over the agent categorization process, and improvements to such process are out of the scope of our work.     

\input{tables/trajectory_analysis_compressed}

\begin{figure*}[ht!]
    \centering
    \includegraphics[width=0.98\textwidth]{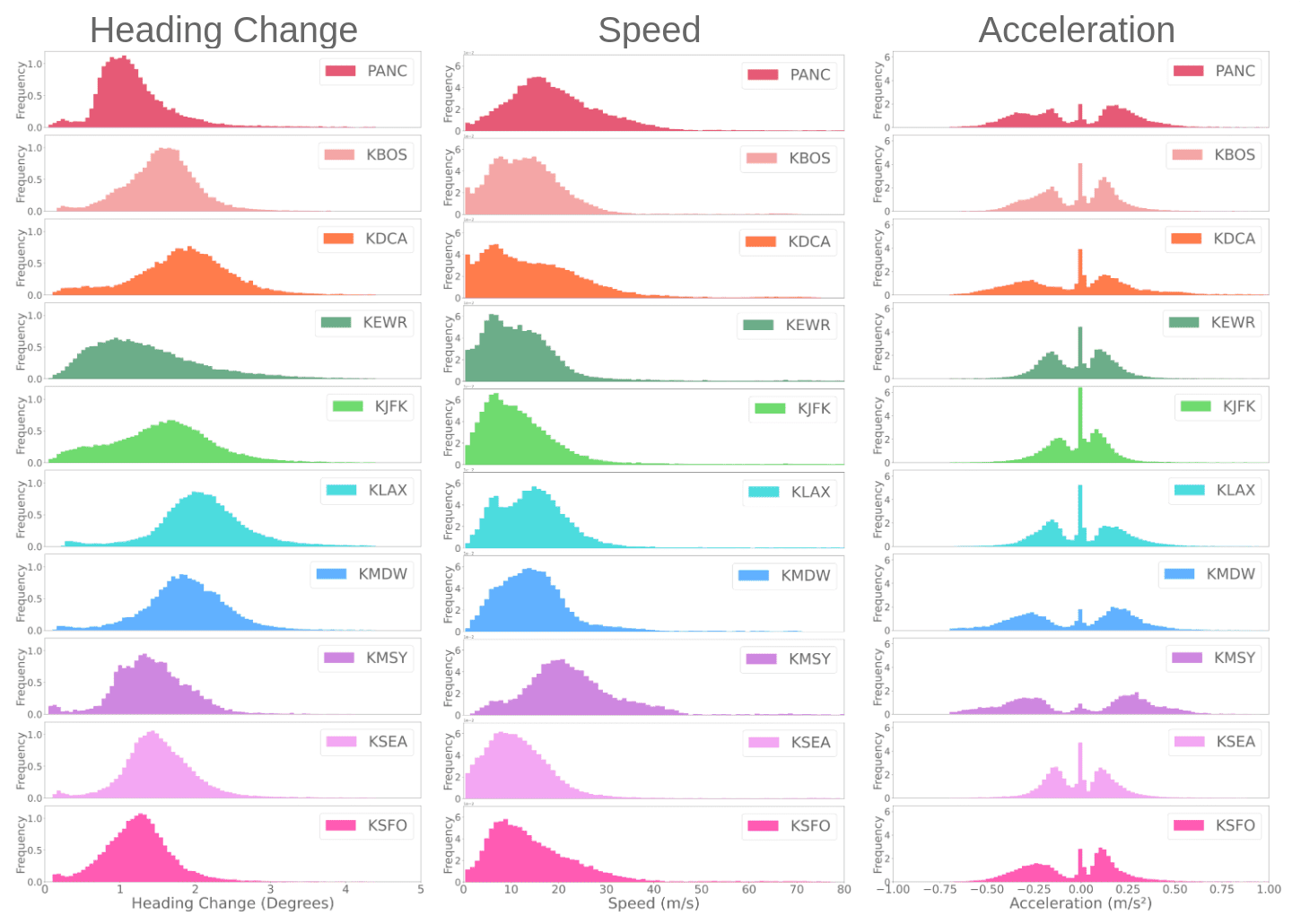}
    \caption{Aircraft motion profiles, showing a distribution shift across different airports. 
    }
    \label{fig:statistic_summary}
\end{figure*}

Finally, we validate aircraft motion profiles across airports in \Cref{fig:statistic_summary}. Specifically, we focus on (1) heading change in subsequent timesteps, (2) mean speed and (3) mean acceleration per trajectory. We see that speed and heading have an ample spread, reflecting the different aircraft behaviors, \exempli take-off taxiing, vacating, etc. The acceleration profile shows a majority of stationary sequences (0 $m/s^2$ bin), which aligns with the behavior of stationary aircraft awaiting clearance or gate assignment, while the noticeable high deceleration and acceleration profiles correspond to aircraft landing and taking off. While diverse across airports, the speed, acceleration, and heading change follow a similar trend, indicating consistent aircraft behavior and verifying that \ameliadataset~accurately captures desired ground operation behaviors.


\input{tables/map_analysis}

\subsection{Map Data Analysis.} \label[appendix]{assec:bench_map_analysis}
 
\Cref{tab:map_analysis} shows that our selected airports also vary at the topological level. First, our selection includes airports with different runway configurations, \exempli \textit{parallel} (P) and \textit{intersecting} (I) runways, and also considering complex topologies which include various \textit{combinations (C)} of both types. We also summarize each airport's graph. We show the \textit{total} and \textit{per-category} number of nodes in each graph. From these numbers, we can contrast the relative complexity between maps, \exempli a \textit{low}-complexity map such as \kmsy~which only has $\sim$1.5k nodes vs. a \textit{high}-complexity one like \klax~with $\sim$11k.

\subsection{Experiment Family Details} \label[appendix]{assec:bench_exp_family}

We consider topological diversity and the crowdedness scale to determine the \multiairport~order. For topological diversity, we focus on the runway organization (see \Cref{tab:map_analysis}), and we use the number of unique data points as a proxy for the crowdedness scale level (see \Cref{tab:trajectory_analysis}). In the list below, we summarize the selection per experiment, \idest the airports that get removed from the training set. In parentheses, we include the airport's runway topology (P: parallel, I: intersecting, and C: combined) and scale rank from 1 (lowest) to 10 (highest).
\begin{itemize}
    \item \seen{10}: We include all airports, \idest all topology types and scales. Here, 6 airports have combined topologies, 2 intersecting runways only, and 2 parallel runways only. 
    \item \seen{7}: We remove one airport per topology type from \seen{10}: \klax~(P, 9), \kmsy~(I, 1), \kjfk~(C, 10).
    \item \seen{4}: We remove one airport per topology type from \seen{7}: \ksea~(P, 8), \kdca~(I, 3), \panc~(C, 2). All of the remaining airports have combined topology and a variety of scales to ensure interaction diversity.
    \item \seen{3}: We remove the largest-scale airport from \seen{4}, \ksfo~(C, 7). 
    \item \seen{2}: We remove the smallest-scale airport from \seen{3}, \kbos~(C, 4).
    \item \seen{1}: We remove the largest-scale airport from \seen{2}, \kewr~(C, 6). The only seen airport is \kmdw~(C, 5).
\end{itemize}

%% file: tables/trajectory_analysis_compressed.tex
\begin{table*}[h!]
\caption{Raw trajectory data summary per airport showing the \textit{total} and \textit{per-category} number of unique agents and data points along with the percentage of interpolated points. }
\label{tab:trajectory_analysis}
\resizebox{0.98\textwidth}{!}{
\begin{tabular}{cccccccccc}
\toprule
\multirow{3}{*}{\textbf{Airport}} &
\multirow{3}{*}{\textbf{Month}} &
\multicolumn{2}{c}{\textbf{Total}} &
\multicolumn{2}{c}{\textbf{Aircraft}} &
\multicolumn{2}{c}{\textbf{Vehicles}} &
\multicolumn{2}{c}{\textbf{Unkown}} \\
\cmidrule(l){3-4} \cmidrule(l){5-6} \cmidrule(l){7-8} \cmidrule(l){9-10}
& 
&
Num. &
Num. Data Points &
Num. Agents &
Num. Data Points &
Num. Agents &
Num. Data Points &
Num. Agents &
Num. Data Points \\
&
&
Agents &
\small{(\% Interp.)} &
\small{(\% \wrt Total)} &
\small{(\% Interp.)} & 
\small{(\% \wrt Total)} &
\small{(\% Interp.)} &
\small{(\% \wrt Total)} &
\small{(\% Interp.)} \\
\midrule
\panc &
    Nov &
    131.35k &
    15.43M (0.22\%) &
    16.96k (12.91\%) &
    5.88M (0.15\%) &
    4.01k (3.12\%) &
    2.51M (0.84\%) &
    110.29k (83.97\%) &
    7.035M (0.06\%) \\
\kbos &
  Jan &
  94.99k &
  23.39M (0.62\%) &
  33.44k (35.20\%) &
  14.77M (0.45\%) &
  4.27k (4.50\%) &
  2.94M (1.33\%) &
  57.29k (60.30\%) &
  5.68M (0.68\%) \\
\kewr &
  Mar &
  140.81k &
  30.86M (4.86\%) &
  38.86k (27.59\%) &
  20.81M (5.67\%) &
  2.21k (1.57\%) &
  1.23M (8.32\%) &
  99.75k (70.84\%) &
  8.81M (2.47\%) \\
\kdca &
  Dec &
  61.85k &
  17.78M (9.94\%) &
  25.81k (41.73\%) &
  8.22M (8.29\%) &
  1.30k (2.10\%) &
  1.00M (16.82\%) &
  34.74k (56.17\%) &
  8.56M (10.72\%) \\
\kjfk &
  Apr &
  645.41k &
  52.82M (9.89\%) &
  45.15k (6.99\%) &
  28.02M (12.03\%) &
  156 (0.02\%) &
  66.31k (4.66\%) &
  600.11k (92.98\%) &
  24.73M (7.47\%) \\
\klax &
  May &
  451.80k &
  42.81M (0.49\%) &
  60.24k (13.33\%) &
  24.67M (0.30\%) &
  14.92k (3.30\%) &
  2.79M (3.33\%) &
  376.70k (83.38\%) &
  15.35M (0.27\%) \\
\kmdw &
  Jun &
  211.92k &
  28.51M (0.84\%) &
  22.47k (10.61\%) &
  6.69M (0.32\%) &
  6.28k (2.97\%) &
  6.15M (3.35\%) &
  183.22k (86.46\%) &
  15.67M (0.07\%) \\
\kmsy &
  Jul &
  49.35k &
  10.37M (0.13\%) &
  10.42k (21.12\%) &
  2.57M (0.08\%) &
  54 (0.11\%) &
  39.89k (3.07\%) &
  38.87k (78.77\%) &
  7.75M (0.13\%) \\
\ksea &
  Aug &
  378.54k &
  38.48M (0.23\%) &
  50.08k (13.23\%) &
  26.13M (0.19\%) &
  290 (0.08\%) &
  274.94k (1.02\%) &
  328.17k (86.70\%) &
  12.077M (0.31\%) \\
\ksfo &
  Sep &
  185.87k &
  36.01M (2.91\%) &
  37.36k (20.10\%) &
  16.67M (2.84\%) &
  14.57k (7.84\%) &
  5.65M (3.26\%) &
  133.94k (72.06\%) &
  13.69M (2.87\%) \\
\midrule
\rowcolor{Gray!10}
\textbf{Total} &
   &
  2,351,883 &
  296,448,313 (3.46\%) &
  340,791 (14.49\%) &
  154,436,226 (3.84\%) &
  48,149 (2.05\%) &
  22,653,655 (3.63\%) &
  1,963,086 (83.47\%) &
  119,354,023 (2.95\%) \\
\bottomrule
\end{tabular}
}
\end{table*}

%% file: tables/map_analysis.tex
\begin{table}[ht!]
\centering
\caption{Map data summary per airport showing runway count and topology, \textit{total} and \textit{per-category} node count. Here, \textbf{P}: parallel, \textbf{I}: intersecting, \textbf{C}: combined runway topology. }
\label{tab:map_analysis}
\resizebox{0.70\textwidth}{!}{
\begin{tabular}{ccccccccc}
\toprule
\multirow{2}{*}{\textbf{Airport}} & \textbf{Num.} & \textbf{Runway} & & \multicolumn{4}{c}{\textbf{Graph Nodes}} \\
\cmidrule{5-8} 
     & \textbf{Runways}  & \textbf{Topology} & & Total  & Runway & Taxiways & Hold Line \\
\midrule
\panc & 3 & C & & 8,016  & 676    & 7,134    & 206    \\
\kbos & 5 & C & & 5,056  & 944    & 3,826    & 286    \\
\kdca & 3 & I & & 8,154  & 438    & 7,590    & 126    \\
\kewr & 3 & C & & 4,804  & 1,424  & 3,090    & 290    \\
\kjfk & 4 & C & & 7,090  & 762    & 5,820    & 508    \\
\klax & 4 & P & & 10,728 & 766    & 9,224    & 738    \\
\kmdw & 5 & C & & 4,132  & 582    & 3,122    & 428    \\
\kmsy & 2 & I & & 1,408  & 474    & 790      & 144    \\
\ksea & 3 & P & & 4,782  & 540    & 4,000    & 242    \\
\ksfo & 4 & C & & 9,032  & 686    & 7,992    & 354    \\
\bottomrule
\end{tabular}
}
\end{table}

%% file: appendix/5_model.tex
\section{\ameliatf~Supplementary} \label[appendix]{asec:model}

This section provides additional details about our scene representation and analyses to justify our design. We provide model implementation details and prediction visualization for one of our models showcasing generalization capabilities. 

\subsection{Scene Representation} \label{assec:model_scene}

We follow an ego-centric scene representation where all agent trajectories and contexts in a scene are encoded \wrt to a pre-selected \textit{ego-agent}. We aim to characterize \textit{safety-relevant} behaviors and interactions within airport operations such as aircraft preparing for take-off, stationary aircraft near hold-short lines, aircraft landing, and taxiing out. To streamline our focus on critical interactions, we developed a selection method inspired by previous work \cite{stoler2023safeshift, glasmacher2022automated}, which identifies relevant ego-agents and produces complex scene representations.

\begin{figure*}[ht!]
    \centering
    \includegraphics[width=0.98\textwidth]{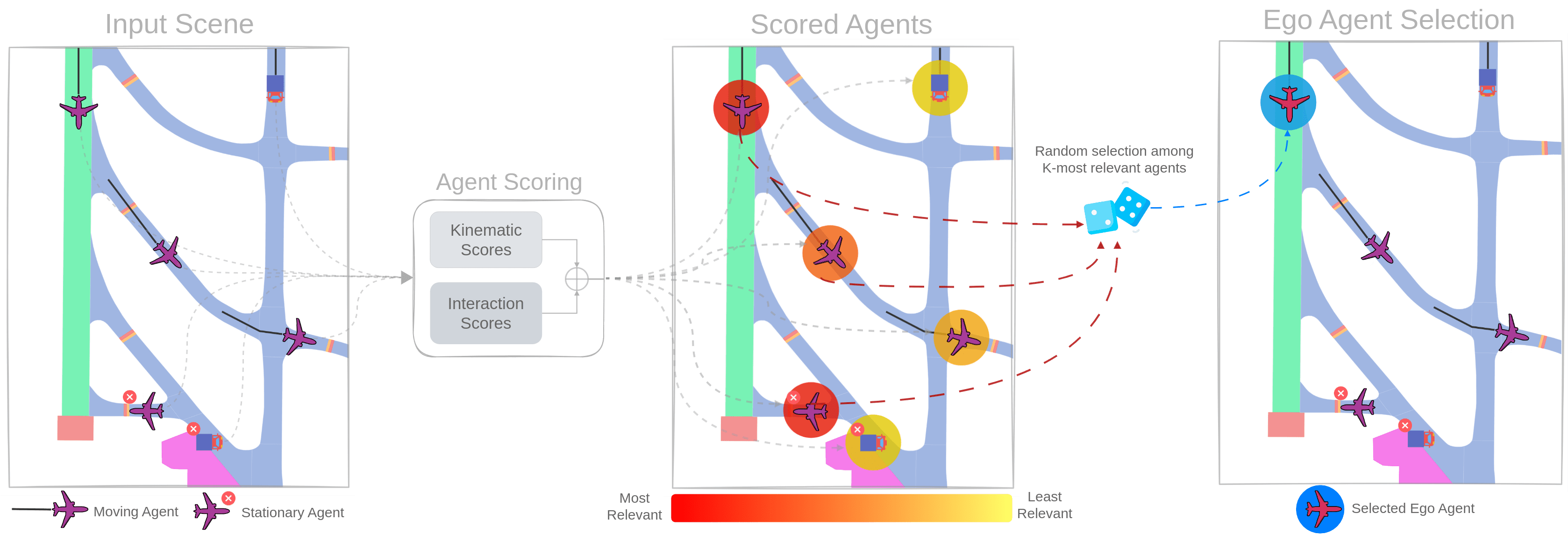}
    \caption{Overview of our scene representation methodology. Given an input scene consisting of $N$ agents, we compute a \textit{kinematic} and \textit{interactive} score for each agent. To represent the final scene, we keep the $K$-highest scored agents, and randomly select an ego-agent among them.  }
    \label{fig:ego_selection}%
\end{figure*}

An overview and toy example of our scene representation and ego-agent selection methodology is shown in \Cref{fig:ego_selection}. The example depicts a scene with a variety of behaviors: an aircraft landing on a \runway, two aircraft vacating a \taxiway, an aircraft waiting near a \holdline, and two vehicles. To select the most relevant agents, our method takes each agent's trajectory and map information and computes a score indicating how \textit{critical} it is; where the higher the score, the more critical the agent is deemed. Our score, specifically, is the result of a function describing each agent's \textit{kinematic} and \textit{interaction} profiles as in \cite{stoler2023safeshift}.


We then select the $K$-most critical agents to represent the final scene. Next, we randomly choose the ego-agent, $e$, among the selected agents and transform the scene \wrt its pose at the last observed timestep, $\state^e_{t_o}$. The final transformed trajectories, $\state_{t_o-H:t_o} \in \R^{K \times H \times D}$, and corresponding historical context, $\context_{t_o-H:t_o} \in \R^{K \times H \times P \times C}$, are encoded using two MLP, $\phi^\state$ and $\phi^\context$, respectively.

\subsection{Implementation Details} \label[appendix]{assec:implementation}


\paragraph{Computational Details.} We trained all of our models on shared servers equipped with an AMD Ryzen Threadripper PRO 5955WX CPU and a single NVIDIA 4090 RTX GPU. 

\paragraph{Scene Representation.} As explained earlier, we consider $T=60$ seconds scenes to capture long-horizon interactions, where the first $H=10$ seconds are the historical segment and the remaining $F=50$ seconds are for prediction to encourage long-horizon and preemptive reasoning from minimal historical information. Agents whose trajectories are shorter than $T$, \exempli agents that exit the scene earlier, or enter it later,  are padded with zeroes. The scene length selection was determined empirically, but as we will show in our scene visualizations, it is sufficient to capture long-horizon and diverse maneuvers. 

\paragraph{Agent Selection.} For the \textbf{kinematic} state, we encode the agent's speed, acceleration, and jerk profiles. We also account for stationary and slow-moving agents near hold-short lines, which are regions where an aircraft should stop before entering a runway. These are especially important since they help ensure an aircraft's proper positioning. To characterize these situations, we incorporate a waiting interval measure weighted by the inverse of the distance to the hold line. For the \textbf{interaction} state, we consider each agent pair and compute a loss of separation value as in \cite{navarro2023sorts}, their minimum time to the conflict point as in~\cite{stoler2023safeshift}.

For a given scene, we select a maximum of $K=5$ agents as the \textit{relevant} agents following the selection process described in \Cref{assec:model_scene}. We obtain each selected agent's local context consisting of the $P=100$ closest points to their location at the last observed timestep. Through this scheme, we obtain over \textbf{19.2M scenes}, comprising \textbf{$\sim$158.2M sequences} or \textbf{9.4B tokens}, where each token represents an agent's state at a given timestep.  

\paragraph{Agent Feature Extractor.} The final transformed trajectories are encoded using a 3-layer MLP with positional embedding.

\paragraph{Context Feature Extractor.} We follow \textit{VectorNet} \cite{gao2020vectornet} to encode map information. Our model encodes individual local patches for each agent in the scene, in contrast to prior works where a shared global context is used for all agents in the scene \cite{shi2022motion, gao2020vectornet, ngiam2021scene}. The intuition behind it is that, in aviation, relevant interactions are not necessarily local as would often be the case for settings such as autonomous driving or navigation in human crowds. For instance, while an autonomous vehicle might be more interested in potential interactions with vehicles at a nearby intersection or conflict region, an aircraft might be interested in the potential interaction with an agent at the other end of the runway. In our domain, however, encoding the global airport map information in the scene may make it difficult for the model to generalize to unseen settings. Thus, we opt for a local context encoding and assume that relative distances between agents and context regions are preserved in the ego-centric scene transformation. 

\paragraph{Scene Encoder.} Our trajectory forecasting model follows a transformer-based architecture that uses a factorized attention scheme for efficient computations as in \cite{ngiam2021scene}. The model maintains a $K \times F \times D$ representation across all transformer layers while interleaving two types of self-attention layers: a \textit{temporal} and an \textit{interaction} one, as well as a \textit{context} cross-attention layer as shown in \Cref{fig:model}. For brevity, we refer to a block comprising these three layers subsequently as TIC. 

The \textbf{temporal} layer is designed as a causal transformer \cite{melnychuk2022causal} used for learning the time-wise dependencies within trajectories. It does so by attending to past features across the time dimension using a mask operator, $\mathbf{M}$, that only \textit{looks} at the previous states from a given timestep. Then, the \textbf{interaction} layer is designed to learn agent-to-agent relationships by attending over the features across the agent dimension. 
We implement both of these layers as multi-head attention (MHA) blocks \cite{vaswani2017attention} which receive the joint feature from the previous layer, as the query, key, and value parameters. 

The \textbf{context} cross-attention layer is then used to exploit the relationships between agent and map features. This layer is also implemented as a MHA block, however, using agents' joint feature as the query input, while the keys and values are obtained from the embedded context. 

\paragraph{Trajectory Decoder.} To model the distribution of possible futures for a scene, we adopt a Gaussian Mixture Model (GMM) following \cite{shi2022motion, chai2019multipath} modeled as a 2-layer MLP that receives the output feature at the final layer of the scene encoder and produces an output representing the agents' predicted means and covariances with an associated likelihood for the $F$ future timesteps. 

\paragraph{Model details.} We use a 5-layer MLP with layer normalization to extract agent features and a 4-layer MLP with batch normalization to extract context features. Our scene encoder is built using 3 TIC blocks that sequentially combine. Each layer in the block consists of an MHA module with $h=8$ attention heads and a 5-layer MLP that outputs the final feature embeddings. Our GMM is built as a 2-layer MLP with a GELU activation and produces $M=4$ output heads. The hidden dimensions of all MLPs are of size $D=256$. The final model comprises \textbf{$\sim$89.8M parameters}. 
We train it for at most 100 epochs 
and use a standard Adam optimizer with a learning rate of $lr=1e-4$. Full details are shown in \Cref{tab:arch}. 

\input{tables/arch}

\subsection{Trajectory Forecasting Results} \label[appendix]{assec:results_forecasting}

\begin{figure*}[ht!]
    \centering
    \includegraphics[width=0.98\textwidth]{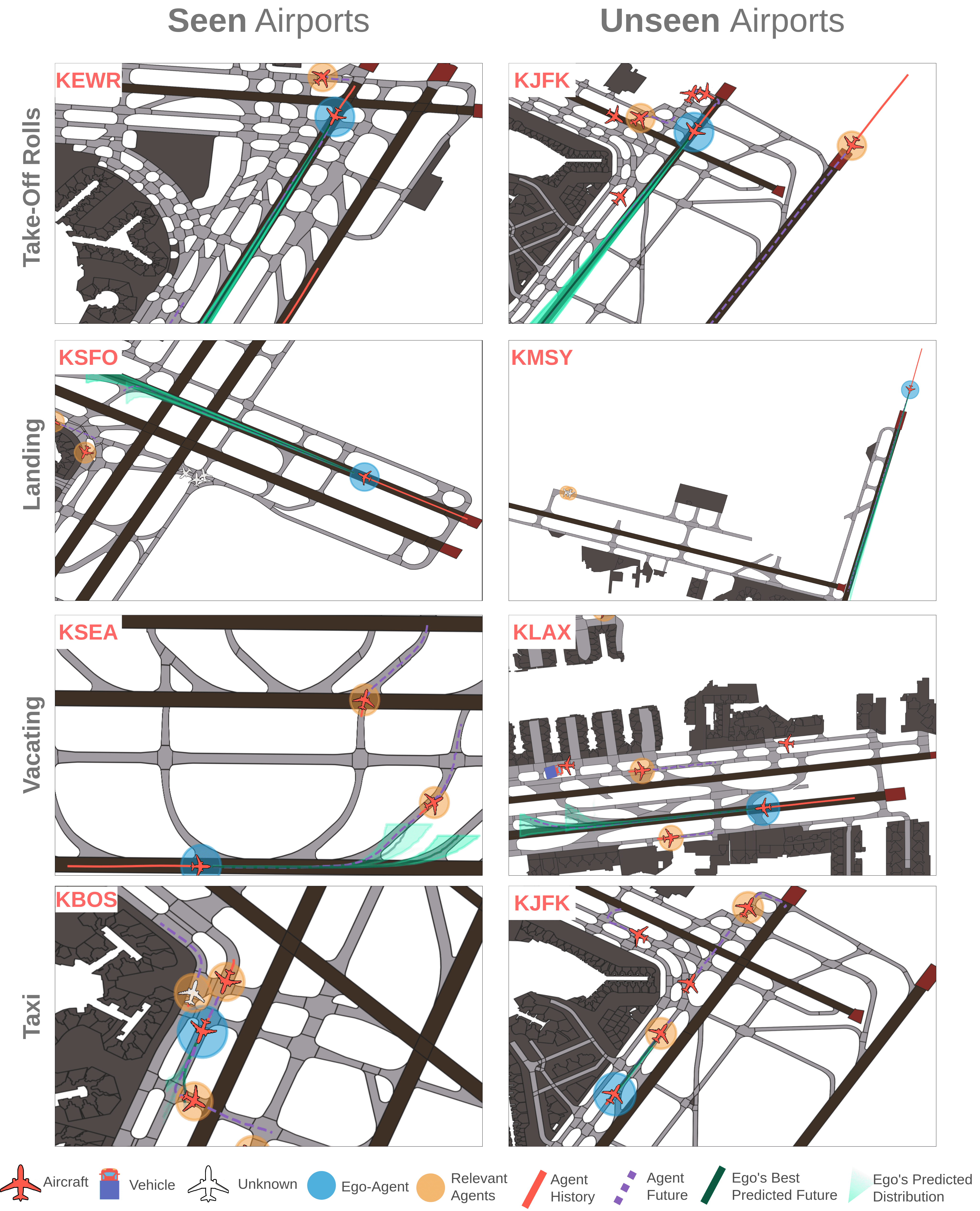}
    \caption{Prediction results for the \seen{7} experiment on four different tasks: take-off roll, landing, vacating a runway, and taxiing. We compare long-horizon ($F=50$s) predictions for \textit{seen} (left) and \textit{unseen} airports (right).}
    \label{fig:generalization_results}%
    \vspace{-0.4cm}%
\end{figure*}

\Cref{fig:generalization_results} depicts qualitative examples for \seen{7}, showing \textit{seen} (left column) and \textit{unseen} (right column) airports. In particular, the figure shows the model's predictions on four common aircraft operations: take-off roll (first row), landing (second), vacating a runway upon landing (third), and taxiing in preparation for take-off (fourth). Through varying airport topologies and agent distributions within the map, these results show that \ameliatf~is able to make strong long-term predictions for each of the tasks in both settings. 

\subsection{Scenario Representation Strategy Validation} \label[appendix]{assec:results_scene}

We study whether our scene selection strategy (\texttt{Critical}) produces more complex and critical interactions by comparing it against a baseline selection approach in which $K$ agents are randomly chosen to represent a scene (\texttt{Random}). As proxies for these components, we analyze the percentage of stationary agents selected as ego agents, as well as the average closest distance between a given agent in a scene and the closest conflict point. 

\input{tables/ego_strategy_scene}

\Cref{tab:ego_strategy_scene} summarizes the per-airport ego-agent selection statistics for both strategies. Each column in \texttt{Critical}, shows in parentheses the average percentage change \wrt \texttt{Random}. Overall, we show that \texttt{Critical} generally selects agents with more dynamic profiles as depicted by the \textbf{15\%} average reduction for the percentage of selected stationary ego agents. The table further shows that our strategy selects more agents on average \textbf{10\%} closer to conflict regions.

\begin{figure*}[t]
    \centering
    \includegraphics[width=0.9\textwidth]{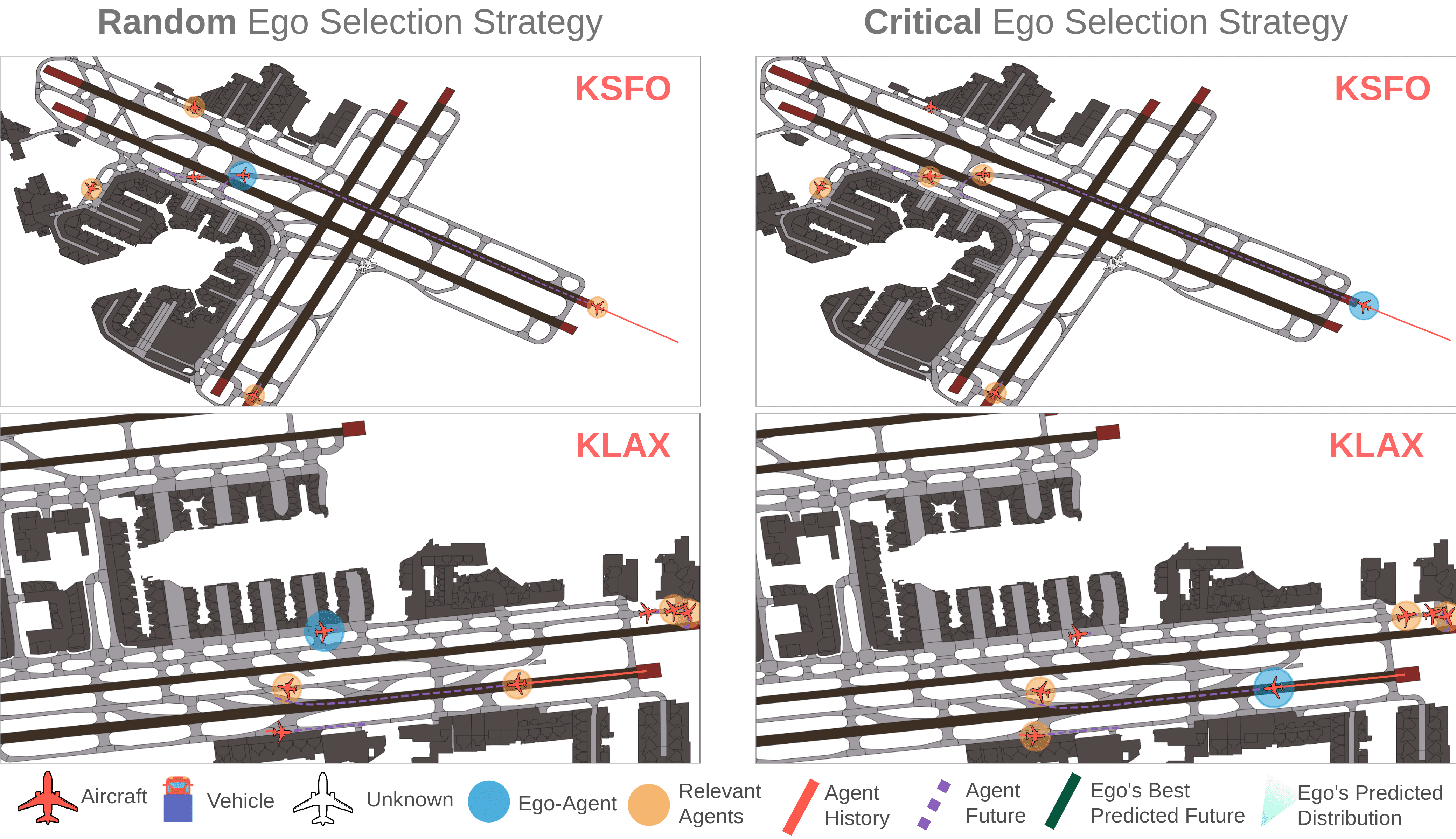}
    \caption{Scene representation and prediction results for the \texttt{Random} and \texttt{Critical} scene representation strategies, showing that the proposed strategy generally selects agents with a more relevant kinematic and interactive profile.}
    \label{fig:ego_results}%
\end{figure*}

We compare qualitative results in \Cref{fig:ego_results}. We show paired scenes for various airports depicting all agents in the scene, the (K-1)-relevant agents (orange halo), and the selected ego agent (blue halo). This figure shows that our selection strategy often selects more dynamic agents, \exempli aircraft preparing for take-off or landing, and agents in conflicting regions, \exempli aircraft vacating a taxiway. Moreover, the agent-to-agent relationships between the selected relevant agents within our strategy are generally more critical. For example, our strategy often selects aircraft that are either vacating or potentially crossing through a runway soon to be used by another aircraft preparing for take-off. In contrast, the random strategy more often selects less dynamic, non-critical agents.

%% file: tables/arch.tex
\begin{table}[!htbp]
\caption{Architecture of \amelia, inspired by \cite{gao2020vectornet, ngiam2021scene}. Here $T=60$, $K=5$, $D_\state=4$, $D_\context=7$, $D_e=256$, $D_f=7$.}
\label{tab:arch}
\resizebox{0.98\textwidth}{!}{
\begin{tabular}{ccccccc}
\toprule
\textbf{Module} & \textbf{Input(s)} & \textbf{Output} & \textbf{Layer Type} & \textbf{Operation Description} & \textbf{Output Dimensions} & \textbf{Num. Parameters} \\
\midrule
\multirow{4}{*}{Agent Feature Extractor $\phi^\state$}
  & $\state$ & $a_1$          & Linear     & -- & $K \times T \times D_{\state} $ & 1.3K \\
  & $a_1$    & $a_2$          & LayerNorm  & -- & $K \times T \times D_{e} $      & 512 \\
  & $a_2$    & $a_3$          & ReLU       & -- & $K \times T \times D_{e} $      & -- \\
  & $a_3$    & $\bar{\state}$ & Linear     & -- & $K \times T \times D_{e} $      & 65.8K \\
  &          & PE & Embedding & Time embedding & $K \times T \times D_{e} $   & 15.4K \\
  &  $\bar{\state}$ + PE  & $\bar{\state}_{\text{PE}}$ & Linear     & -- & $K \times T \times D_{e} $      & 65.8K \\
\midrule
\multirow{10}{*}{Context Feature Extractor $\phi^\context$}
  & $\context$ & $c_1$ & Linear     & -- & $K \times P \times D_{\context} $ & 1.5K \\
  & $c_1$      & $c_2$ & BatchNorm  & -- & $K \times P \times D_{e} $        & 200  \\
  & $c_2$      & $c_3$ & ReLU       & -- & $K \times P \times D_{e} $        & --  \\
  & $c_3$      & $c_4$ & Linear     & -- & $K \times P \times D_{e} $        & 65.8K \\
  & $c_4$      & $c_5$ & BatchNorm  & -- & $K \times P \times D_{e} $        & 200  \\
  & $c_5$      & $c_6$ & ReLU       & -- & $K \times P \times D_{e} $        & --  \\
  & $c_6$      & $c_7$ & Linear     & -- & $K \times P \times D_{e} $        & 81.9K \\
  & $c_7$      & $c_8$ & BatchNorm  & -- & $K \times P \times D_{e} $        & 6.4K  \\
  & $c_8$      & $c_9$ & ReLU       & -- & $K \times P \times D_{e} $        & --  \\
  & $c_9$      & $\bar{\context}$ & Linear & Feature down-projection & $K \times T \times D_{e} $ & 819K \\
\midrule
\multirow{9}{*}{Scene Encoder}
  & $\bar{\state}_{\text{PE}}$ & $e_1$ & Transformer & Self-attention across $T$ & $K \times T \times D_{e}$ & 789K \\
  & $e_1$ & $e_2$ & Transformer & Self-attention across $K$ & $K \times T \times D_{e} $ & 789K \\
  & $e_2$, $\bar{\context}$ & $e_3$ & Transformer & Cross-attention & $K \times T \times D_{e} $ & 789K \\
  & $e_3$ & $e_4$ & Transformer & Self-attention across $T$ & $K \times T \times D_{e}$ & 789K \\
  & $e_4$ & $e_5$ & Transformer & Self-attention across $K$ & $K \times T \times D_{e} $ & 789K \\
  & $e_5$, $\bar{\context}$ & $e_6$ & Transformer & Cross-attention & $K \times T \times D_{e} $ & 789K \\
  & $e_6$ & $e_7$ & Transformer & Self-attention across $T$ & $K \times T \times D_{e}$ & 789K \\
  & $e_7$ & $e_8$ & Transformer & Self-attention across $K$ & $K \times T \times D_{e} $ & 789K \\
  & $e_8$ & $e_9$ & Linear & Refinement Layer & $K \times T \times D_{e} $ & 525K \\
\midrule
\multirow{4}{*}{Scene Decoder}
  & $e_9$ & $d_1$ & Linear & -- & $M \times K \times T \times D_{e}$ & 16.6K\\
  & $d_1$ & $d_2$ & GeLU   & -- & $M \times K \times T \times D_{e}$ & --   \\
  & $d_2$ & $\mu_\state, \Sigma_\state, \rho$ & Linear & -- & $M \times K \times T \times D_{f}$ & 1.8K \\
  & $\rho$ & $\rho$ & Softmax & Softmax across M & $M \times K \times T \times 1$ & --  \\
\bottomrule
\end{tabular}
}
\end{table}

%% file: tables/ego_strategy_scene.tex
\begin{table*}[ht!]
\caption{Ego-agent selection statistics for the \texttt{Random} and \texttt{Critical} strategies. In parentheses, we show the relative percentage difference between relevant and random strategies.}
\resizebox{0.98\textwidth}{!}{
\begin{tabular}{ccccccccccccc}
\toprule
 & & & \multicolumn{4}{c}{\textbf{\texttt{Random} Selection Strategy}} & & & \multicolumn{4}{c}{\textbf{\texttt{Critical} Selection Strategy}} \\
\cmidrule(l){4-8}\cmidrule(l){10-13}
\multirow{2}{*}{\textbf{Airport}} & Total Num. & & Stationary & \multicolumn{3}{c}{Avg. Closest Dist. to Conflict Point (m)} & & & Stationary & \multicolumn{3}{c}{Avg. Closest Dist. to Conflict Point (m)} \\
\cmidrule(l){5-7}\cmidrule(l){11-13}
& Ego-agents & & Ego-agents (\%) & All Agents & Stationary Agents & Moving Agents & & & Ego-agents (\%) & All Agents & Stationary Agents & Moving Agents \\
\midrule
\panc
& 481,353 & & 1.84 & 74.44 & 74.15 & 107.87 &
& & 1.75 (\textcolor{PineGreen!50}{-4.89\%}) & 72.26 (\textcolor{PineGreen!50}{-2.93\%}) 
& 72.08 (\textcolor{PineGreen!50}{-2.80\%})	& 96.40	(\textcolor{PineGreen!50}{-10.64\%}) \\
\kbos 
& 435,162 & & 15.28 & 82.30 & 69.43 & 153.66 & 
& & 11.30 (\textcolor{PineGreen}{-26.05\%}) & 	70.44 (\textcolor{PineGreen!80}{-14.42\%}) 
& 60.44	(\textcolor{PineGreen!80}{-12.95\%}) & 148.96 (\textcolor{PineGreen!50}{-3.06\%}) \\
\kdca 
& 455,366 & & 43.92 & 103.48 & 95.84 & 113.25 & 
& & 36.84 (\textcolor{PineGreen!90}{-16.12\%}) & 99.61 (\textcolor{PineGreen!50}{-3.74\%}) 
&  92.25 (\textcolor{PineGreen!50}{-3.74\%}) & 112.23 (\textcolor{PineGreen!40}{-0.90\%}) \\
\kewr 
& 195,615 & & 10.58 & 95.14 &	92.95 &	113.62 &
& & 7.97 (\textcolor{PineGreen}{-24.67\%}) & 83.29 (\textcolor{PineGreen!90}{-12.45\%}) 
& 81.40 (\textcolor{PineGreen!90}{-12.43\%}) & 105.12 (\textcolor{PineGreen!90}{-7.49\%}) \\
\kjfk 
& 54,458 & & 38.29 & 88.30 & 78.38 & 104.27 &
& & 30.60 (\textcolor{PineGreen}{-20.08\%}) & 74.79 (\textcolor{PineGreen!90}{-15.30\%}) 
& 	68.42 (\textcolor{PineGreen!80}{-12.71\%}) & 89.23 (\textcolor{PineGreen!80}{-14.43\%}) \\ 
\klax 
& 125,602 & & 5.26 & 50.55 & 49.24 & 74.04 & 
& & 4.27 (\textcolor{PineGreen!90}{-18.82\%}) & 47.01 (\textcolor{PineGreen!70}{-6.99\%}) 
& 46.52	(\textcolor{PineGreen!70}{-5.53\%}) & 58.15 (\textcolor{PineGreen}{-21.47\%}) \\ 
\kmdw 
& 268,734 & & 9.49 & 66.58 & 70.54 & 28.84 & 
& & 9.91 (\textcolor{Red!70}{4.43\%}) & 58.14 (\textcolor{PineGreen!80}{-12.69\%}) 
& 61.93	(\textcolor{PineGreen!80}{-12.21\%}) & 23.69 (\textcolor{PineGreen!90}{-17.86\%}) \\ 
\kmsy 
& 673,439 & & 17.69 &	147.63 & 149.73 & 137.85 & 
& & 17.87 (\textcolor{Red!50}{1.02\%}) & 144.20	(\textcolor{PineGreen!50}{-2.32\%}) 
& 146.11 (\textcolor{PineGreen!50}{-2.42\%}) & 135.41	(\textcolor{PineGreen!50}{-1.77\%}) \\ 
\ksea 
& 85,948 & & 4.51 & 81.19 & 79.17 & 124.06 & 
& & 3.24 (\textcolor{PineGreen}{-28.16\%}) & 77.57 (\textcolor{PineGreen!50}{-4.45\%}) 
& 76.15 (\textcolor{PineGreen!50}{-3.81\%}) & 120.03 (\textcolor{PineGreen!50}{-3.25\%}) \\
\ksfo 
& 186,436 & & 26.22 & 82.76 & 78.19 & 95.62 & 
& & 20.70 (\textcolor{PineGreen}{-21.05\%}) & 68.25 (\textcolor{PineGreen!90}{-17.54\%}) 
& 65.34	(\textcolor{PineGreen!90}{-16.44\%}) & 79.42 (\textcolor{PineGreen!90}{-16.95\%}) \\ 
\midrule
\rowcolor{Gray!10}
\textbf{Average} 
& 296,211 & & 17.31 & 87.24 & 83.76 & 105.31 & 
& & 14.45 (-15.44\%) & 79.55 (-9.28\%) 
& 77.06	(-8.50\%) & 96.86 (-9.78\%) \\ 
\bottomrule
\end{tabular}
}
\scriptsize{
\begin{tablenotes}
\item \textbf{Note:} The percentage of stationary agents and the average closest distance to a conflict point are reported for the test set. 
\end{tablenotes}
}
\label{tab:ego_strategy_scene}
\end{table*}

%% file: main.bbl
\begin{thebibliography}{57}
\newcommand{\enquote}[1]{``#1''}
\providecommand{\natexlab}[1]{#1}
\providecommand{\url}[1]{\texttt{#1}}
\providecommand{\urlprefix}{URL }
\expandafter\ifx\csname urlstyle\endcsname\relax
  \providecommand{\doi}[1]{\discretionary{}{}{}https://doi.org/#1}\else
  \providecommand{\doi}[1]{\discretionary{}{}{}\urlstyle{rm}\url{https://doi.org/#1}}\fi

\bibitem[{Daugherty(2023)}]{daugherty2023we}
Daugherty, A., \enquote{‘We were very lucky’: Near-collisions spark new worries for air travel,} \emph{POLITICO}, 2023.

\bibitem[{Leussink et~al.(2024)Leussink, Kaneko, and Barrington}]{leussink2024japan}
Leussink, D., Kaneko, K., and Barrington, L., \enquote{Japan Airlines counts losses from wrecked Tokyo Plane,} \emph{Reuters}, 2024.

\bibitem[{Helmuth~Rosales and Ismay(2025)}]{helmuth2025what}
Helmuth~Rosales, M.~G., K.K. Rebecca~Lai, and Ismay, J., \enquote{What the Black Hawk Pilots Could See, Just Before the Crash,} \emph{The New York Times}, 2025.

\bibitem[{Ruberg(2025)}]{ruberg2025one}
Ruberg, S., \enquote{1 Killed as Plane Owned by Mötley Crüe Singer Strikes Parked Jet,} \emph{The New York Times}, 2025.

\bibitem[{Ember and Steel(2023)}]{EmberSteel2023}
Ember, S., and Steel, E., \enquote{Airline Close Calls Happen Far More Often Than Previously Known,} \emph{The New York Times}, 2023.

\bibitem[{Krolik(2025)}]{Krolik2025}
Krolik, A., \enquote{285 of 313 Air Traffic Control Facilities Are Understaffed,} \emph{The New York Times}, 2025.

\bibitem[{Association(2022)}]{iata2022air}
Association, I. A.~T., \enquote{Air Passenger Market Analysis - December 2022,} Tech. rep., IATA, 2022.

\bibitem[{Aratani(2023)}]{aratoni2023tsa}
Aratani, L., \enquote{TSA broke record on Sunday for number of passengers screened,} \emph{The Washington Post}, 2023.

\bibitem[{Degas et~al.(2022)Degas, Islam, Hurter, Barua, Rahman, Poudel, Ruscio, Ahmed, Begum, Rahman et~al.}]{degas2022survey}
Degas, A., Islam, M.~R., Hurter, C., Barua, S., Rahman, H., Poudel, M., Ruscio, D., Ahmed, M.~U., Begum, S., Rahman, M.~A., et~al., \enquote{A survey on artificial intelligence (ai) and explainable ai in air traffic management: Current trends and development with future research trajectory,} \emph{Applied Sciences}, Vol.~12, No.~3, 2022, p. 1295.

\bibitem[{Sun et~al.(2020)Sun, Kretzschmar, Dotiwalla, Chouard, Patnaik, Tsui, Guo, Zhou, Chai, Caine et~al.}]{sun2020scalability}
Sun, P., Kretzschmar, H., Dotiwalla, X., Chouard, A., Patnaik, V., Tsui, P., Guo, J., Zhou, Y., Chai, Y., Caine, B., et~al., \enquote{Scalability in perception for autonomous driving: Waymo open dataset,} \emph{Proceedings of the IEEE/CVF conference on computer vision and pattern recognition}, 2020, pp. 2446--2454.

\bibitem[{Chang et~al.(2019)Chang, Lambert, Sangkloy, Singh, Bak, Hartnett, Wang, Carr, Lucey, Ramanan, and Hays}]{fang2019argoverse}
Chang, M., Lambert, J., Sangkloy, P., Singh, J., Bak, S., Hartnett, A., Wang, D., Carr, P., Lucey, S., Ramanan, D., and Hays, J., \enquote{Argoverse: 3D Tracking and Forecasting With Rich Maps,} \emph{{IEEE} Conference on Computer Vision and Pattern Recognition, {CVPR} 2019, Long Beach, CA, USA, June 16-20, 2019}, Computer Vision Foundation / {IEEE}, 2019, pp. 8748--8757.
\newblock \doi{10.1109/CVPR.2019.00895}.

\bibitem[{Caesar et~al.(2020)Caesar, Bankiti, Lang, Vora, Liong, Xu, Krishnan, Pan, Baldan, and Beijbom}]{caesar2020nuscenes}
Caesar, H., Bankiti, V., Lang, A.~H., Vora, S., Liong, V.~E., Xu, Q., Krishnan, A., Pan, Y., Baldan, G., and Beijbom, O., \enquote{nuscenes: A multimodal dataset for autonomous driving,} \emph{Proceedings of the IEEE/CVF conference on computer vision and pattern recognition}, 2020, pp. 11621--11631.

\bibitem[{Rudenko et~al.(2020)Rudenko, Palmieri, Herman, Kitani, Gavrila, and Arras}]{rudenko2020humanmotion}
Rudenko, A., Palmieri, L., Herman, M., Kitani, K.~M., Gavrila, D.~M., and Arras, K.~O., \enquote{Human motion trajectory prediction: a survey,} \emph{Int. J. Robotics Res.}, Vol.~39, No.~8, 2020.
\newblock \doi{10.1177/0278364920917446}.

\bibitem[{Pellegrini et~al.(2009)Pellegrini, Ess, Schindler, and Gool}]{pellegrini2009eth}
Pellegrini, S., Ess, A., Schindler, K., and Gool, L.~V., \enquote{You'll never walk alone: Modeling social behavior for multi-target tracking,} \emph{{IEEE} 12th International Conference on Computer Vision, {ICCV} 2009, Kyoto, Japan, September 27 - October 4, 2009}, {IEEE} Computer Society, 2009, pp. 261--268.
\newblock \doi{10.1109/ICCV.2009.5459260}.

\bibitem[{Lerner et~al.(2007)Lerner, Chrysanthou, and Lischinski}]{lerner2007ucy}
Lerner, A., Chrysanthou, Y., and Lischinski, D., \enquote{Crowds by Example,} \emph{Comput. Graph. Forum}, Vol.~26, No.~3, 2007, pp. 655--664.
\newblock \doi{10.1111/j.1467-8659.2007.01089.x}.

\bibitem[{Navarro et~al.(2024{\natexlab{a}})Navarro, Patrikar, Dantas, Baijal, Higgins, Scherer, and Oh}]{navarro2023sorts}
Navarro, I., Patrikar, J., Dantas, J. P.~A., Baijal, R., Higgins, I., Scherer, S., and Oh, J., \enquote{SoRTS: Learned Tree Search for Long Horizon Social Robot Navigation,} \emph{IEEE Robotics and Automation Letters}, Vol.~9, No.~4, 2024{\natexlab{a}}, pp. 3759--3766.
\newblock \doi{10.1109/LRA.2024.3370051}.

\bibitem[{Patrikar et~al.(2022{\natexlab{a}})Patrikar, Dantas, Ghosh, Kapoor, Higgins, Aloor, Navarro, Sun, Stoler, Hamidi et~al.}]{patrikar2022challenges}
Patrikar, J., Dantas, J., Ghosh, S., Kapoor, P., Higgins, I., Aloor, J.~J., Navarro, I., Sun, J., Stoler, B., Hamidi, M., et~al., \enquote{Challenges in close-proximity safe and seamless operation of manned and unmanned aircraft in shared airspace,} \emph{arXiv preprint arXiv:2211.06932}, 2022{\natexlab{a}}.

\bibitem[{Muthali et~al.(2023)Muthali, Shen, Deglurkar, Lim, Roelofs, Faust, and Tomlin}]{muthali2023multi}
Muthali, A., Shen, H., Deglurkar, S., Lim, M.~H., Roelofs, R., Faust, A., and Tomlin, C., \enquote{Multi-agent reachability calibration with conformal prediction,} \emph{arXiv preprint arXiv:2304.00432}, 2023.

\bibitem[{Sui et~al.(2023)Sui, Chen, and Zhou}]{sui2023conflict}
Sui, D., Chen, H., and Zhou, T., \enquote{A Conflict Resolution Strategy at a Taxiway Intersection by Combining a Monte Carlo Tree Search with Prior Knowledge,} \emph{Aerospace}, Vol.~10, No.~11, 2023, p. 914.

\bibitem[{Kong and Mahadevan(2023)}]{kong2023identifying}
Kong, Y., and Mahadevan, S., \enquote{Identifying Anomalous Behavior in Aircraft Landing Trajectory Using a Bayesian Autoencoder,} \emph{Journal of Aerospace Information Systems}, 2023, pp. 1--9.

\bibitem[{Memarzadeh et~al.(2023)Memarzadeh, Matthews, and Weckler}]{memarzadeh2023anomaly}
Memarzadeh, M., Matthews, B.~L., and Weckler, D.~I., \enquote{Anomaly Detection in Flight Operational Data Using Deep Learning,} \emph{System-Wide Safety Technical Challenge 1 Close Out Event}, 2023.

\bibitem[{Liu et~al.(2014)Liu, Hansen, Gupta, Malik, and Jung}]{liu2014predictability}
Liu, Y., Hansen, M., Gupta, G., Malik, W., and Jung, Y., \enquote{Predictability impacts of airport surface automation,} \emph{Transportation Research Part C: Emerging Technologies}, Vol.~44, 2014, pp. 128--145.

\bibitem[{Lee et~al.(2016)Lee, Malik, and Jung}]{lee2016taxi}
Lee, H., Malik, W., and Jung, Y.~C., \enquote{Taxi-out time prediction for departures at Charlotte airport using machine learning techniques,} \emph{16th AIAA Aviation Technology, Integration, and Operations Conference}, 2016, p. 3910.

\bibitem[{Wang et~al.(2023)Wang, Bi, Xie, and Zhao}]{wang2023data}
Wang, F., Bi, J., Xie, D., and Zhao, X., \enquote{A data-driven prediction model for aircraft taxi time by considering time series about gate and real-time factors,} \emph{Transportmetrica A: Transport Science}, Vol.~19, No.~3, 2023, p. 2071353.

\bibitem[{Alsalous and Hotle(2023)}]{alsalous2023deicing}
Alsalous, O., and Hotle, S., \enquote{Deicing Facility Capacity and Delay Estimation Using ASDE-X Data: Chicago O’Hare Simulation Case Study,} \emph{Transportation Research Record}, 2023, p. 03611981231185147.

\bibitem[{Gui et~al.(2021)Gui, Zhang, Peng, and Yang}]{gui2021data}
Gui, X., Zhang, J., Peng, Z., and Yang, C., \enquote{Data-driven method for the prediction of estimated time of arrival,} \emph{Transportation Research Record}, Vol. 2675, No.~12, 2021, pp. 1291--1305.

\bibitem[{Li and Ryerson(2019)}]{li2019reviewing}
Li, M.~Z., and Ryerson, M.~S., \enquote{Reviewing the DATAS of aviation research data: Diversity, availability, tractability, applicability, and sources,} \emph{Journal of Air Transport Management}, Vol.~75, 2019, pp. 111--130.

\bibitem[{Zhang et~al.(2022)Zhang, Zhong, and Mahadevan}]{zhang2022airport}
Zhang, X., Zhong, S., and Mahadevan, S., \enquote{Airport surface movement prediction and safety assessment with spatial--temporal graph convolutional neural network,} \emph{Transportation research part C: emerging technologies}, Vol. 144, 2022, p. 103873.

\bibitem[{Park and Kim(2023)}]{park2023influential}
Park, D.~K., and Kim, J.~K., \enquote{Influential factors to aircraft taxi time in airport,} \emph{Journal of Air Transport Management}, Vol. 106, 2023, p. 102321.

\bibitem[{Ranson(2011)}]{ranson2011faa}
Ranson, L., \enquote{FAA completes roll-out of ASDE-X anti-incursion surveillance,} \emph{Flight International}, Vol. 180, No. 5313, 2011, pp. 15--15.

\bibitem[{Navarro et~al.(2024{\natexlab{b}})Navarro, Ortega, Patrikar, Wang, Ye, Park, Oh, and Scherer}]{navarro2024ameliatf}
Navarro, I., Ortega, P., Patrikar, J., Wang, H., Ye, Z., Park, J.~H., Oh, J., and Scherer, S., \enquote{AmeliaTF: A Large Model and Dataset for Airport Surface Movement Forecasting,} \emph{AIAA AVIATION FORUM AND ASCEND 2024}, 2024{\natexlab{b}}, p. 4251.

\bibitem[{Patrikar et~al.(2022{\natexlab{b}})Patrikar, Moon, Oh, and Scherer}]{patrikar2022predicting}
Patrikar, J., Moon, B., Oh, J., and Scherer, S., \enquote{Predicting like a pilot: Dataset and method to predict socially-aware aircraft trajectories in non-towered terminal airspace,} \emph{2022 International Conference on Robotics and Automation (ICRA)}, IEEE, 2022{\natexlab{b}}, pp. 2525--2531.

\bibitem[{Gao et~al.(2020)Gao, Sun, Zhao, Shen, Anguelov, Li, and Schmid}]{gao2020vectornet}
Gao, J., Sun, C., Zhao, H., Shen, Y., Anguelov, D., Li, C., and Schmid, C., \enquote{Vectornet: Encoding hd maps and agent dynamics from vectorized representation,} \emph{Proceedings of the IEEE/CVF Conference on Computer Vision and Pattern Recognition}, 2020, pp. 11525--11533.

\bibitem[{Nayakanti et~al.(2023)Nayakanti, Al-Rfou, Zhou, Goel, Refaat, and Sapp}]{nayakanti2023wayformer}
Nayakanti, N., Al-Rfou, R., Zhou, A., Goel, K., Refaat, K.~S., and Sapp, B., \enquote{Wayformer: Motion forecasting via simple \& efficient attention networks,} \emph{2023 IEEE International Conference on Robotics and Automation (ICRA)}, IEEE, 2023, pp. 2980--2987.

\bibitem[{Cui et~al.(2023)Cui, Casas, Wong, Suo, and Urtasun}]{cui2023gorela}
Cui, A., Casas, S., Wong, K., Suo, S., and Urtasun, R., \enquote{Gorela: Go relative for viewpoint-invariant motion forecasting,} \emph{2023 IEEE International Conference on Robotics and Automation (ICRA)}, IEEE, 2023, pp. 7801--7807.

\bibitem[{Shi et~al.(2022)Shi, Jiang, Dai, and Schiele}]{shi2022motion}
Shi, S., Jiang, L., Dai, D., and Schiele, B., \enquote{Motion transformer with global intention localization and local movement refinement,} \emph{Advances in Neural Information Processing Systems}, Vol.~35, 2022, pp. 6531--6543.

\bibitem[{Yuan et~al.(2021)Yuan, Weng, Ou, and Kitani}]{yuan2021agentformer}
Yuan, Y., Weng, X., Ou, Y., and Kitani, K.~M., \enquote{Agentformer: Agent-aware transformers for socio-temporal multi-agent forecasting,} \emph{Proceedings of the IEEE/CVF International Conference on Computer Vision}, 2021, pp. 9813--9823.

\bibitem[{Ngiam et~al.(2021)Ngiam, Vasudevan, Caine, Zhang, Chiang, Ling, Roelofs, Bewley, Liu, Venugopal et~al.}]{ngiam2021scene}
Ngiam, J., Vasudevan, V., Caine, B., Zhang, Z., Chiang, H.-T.~L., Ling, J., Roelofs, R., Bewley, A., Liu, C., Venugopal, A., et~al., \enquote{Scene transformer: A unified architecture for predicting future trajectories of multiple agents,} \emph{International Conference on Learning Representations}, 2021, pp. 541--556.

\bibitem[{Navarro and Oh(2022)}]{navarro2022socialpattern}
Navarro, I., and Oh, J., \enquote{Social-PatteRNN: Socially-Aware Trajectory Prediction Guided by Motion Patterns,} \emph{2022 IEEE/RSJ International Conference on Intelligent Robots and Systems (IROS)}, IEEE, 2022, pp. 9859--9864.

\bibitem[{Alahi et~al.(2016)Alahi, Goel, Ramanathan, Robicquet, Fei-Fei, and Savarese}]{alahi2016social}
Alahi, A., Goel, K., Ramanathan, V., Robicquet, A., Fei-Fei, L., and Savarese, S., \enquote{Social lstm: Human trajectory prediction in crowded spaces,} \emph{Proceedings of the IEEE conference on computer vision and pattern recognition}, 2016, pp. 961--971.

\bibitem[{Zhao and Oh(2020)}]{zhao2020noticing}
Zhao, D., and Oh, J., \enquote{Noticing motion patterns: A temporal cnn with a novel convolution operator for human trajectory prediction,} \emph{IEEE Robotics and Automation Letters}, Vol.~6, No.~2, 2020, pp. 628--634.

\bibitem[{Guo et~al.(2023)Guo, Wu, Wu, Zhang, Law, and Lin}]{guo_flightbert_2023}
Guo, D., Wu, E.~Q., Wu, Y., Zhang, J., Law, R., and Lin, Y., \enquote{{FlightBERT}: {Binary} {Encoding} {Representation} for {Flight} {Trajectory} {Prediction},} \emph{IEEE Transactions on Intelligent Transportation Systems}, Vol.~24, No.~2, 2023, pp. 1828--1842.
\newblock \doi{10.1109/TITS.2022.3219923}, \urlprefix\url{https://ieeexplore.ieee.org/document/9945661}, conference Name: IEEE Transactions on Intelligent Transportation Systems.

\bibitem[{Guo et~al.(2024)Guo, Zhang, Yan, Zhang, and Lin}]{guo_flightbert_2024}
Guo, D., Zhang, Z., Yan, Z., Zhang, J., and Lin, Y., \enquote{{FlightBERT}++: {A} {Non}-autoregressive {Multi}-{Horizon} {Flight} {Trajectory} {Prediction} {Framework},} \emph{Proceedings of the AAAI Conference on Artificial Intelligence}, Vol.~38, No.~1, 2024, pp. 127--134.
\newblock \doi{10.1609/aaai.v38i1.27763}, \urlprefix\url{https://ojs.aaai.org/index.php/AAAI/article/view/27763}, number: 1.

\bibitem[{Zhang et~al.(2023)Zhang, Guo, Zhou, Zhang, and Lin}]{zhang_flight_2023}
Zhang, Z., Guo, D., Zhou, S., Zhang, J., and Lin, Y., \enquote{Flight trajectory prediction enabled by time-frequency wavelet transform,} \emph{Nature Communications}, Vol.~14, No.~1, 2023, p. 5258.
\newblock \doi{10.1038/s41467-023-40903-9}, \urlprefix\url{https://www.nature.com/articles/s41467-023-40903-9}, publisher: Nature Publishing Group.

\bibitem[{Yin et~al.(2023)Yin, Zhang, Zhang, Zhang, and Xiang}]{yin2023context}
Yin, Y., Zhang, S., Zhang, Y., Zhang, Y., and Xiang, S., \enquote{Context-aware Aircraft Trajectory Prediction with Diffusion Models,} \emph{2023 IEEE 26th International Conference on Intelligent Transportation Systems (ITSC)}, IEEE, 2023, pp. 5312--5317.

\bibitem[{Ivanovic et~al.(2023)Ivanovic, Song, Gilitschenski, and Pavone}]{ivanovic2023trajdata}
Ivanovic, B., Song, G., Gilitschenski, I., and Pavone, M., \enquote{trajdata: A unified interface to multiple human trajectory datasets,} \emph{Advances in Neural Information Processing Systems}, Vol.~36, 2023, pp. 27582--27593.

\bibitem[{Map(2014)}]{map2014open}
Map, O.~S., \enquote{Open street map,} \emph{Online: https://www. openstreetmap. org. Search in}, 2014.

\bibitem[{Boeing(2017)}]{boeing2017osmnx}
Boeing, G., \enquote{OSMnx: New methods for acquiring, constructing, analyzing, and visualizing complex street networks,} \emph{Computers, Environment and Urban Systems}, Vol.~65, 2017, pp. 126--139.

\bibitem[{Stoler et~al.(2024)Stoler, Navarro, Jana, Hwang, Francis, and Oh}]{stoler2023safeshift}
Stoler, B., Navarro, I., Jana, M., Hwang, S., Francis, J., and Oh, J., \enquote{SafeShift: Safety-Informed Distribution Shifts for Robust Trajectory Prediction in Autonomous Driving,} \emph{2024 IEEE Intelligent Vehicles Symposium (IV)}, 2024, pp. 1179--1186.
\newblock \doi{10.1109/IV55156.2024.10588828}.

\bibitem[{Ho et~al.(2019)Ho, Kalchbrenner, Weissenborn, and Salimans}]{ho2019axial}
Ho, J., Kalchbrenner, N., Weissenborn, D., and Salimans, T., \enquote{Axial attention in multidimensional transformers,} \emph{arXiv preprint arXiv:1912.12180}, 2019.

\bibitem[{Chai et~al.(2019)Chai, Sapp, Bansal, and Anguelov}]{chai2019multipath}
Chai, Y., Sapp, B., Bansal, M., and Anguelov, D., \enquote{Multipath: Multiple probabilistic anchor trajectory hypotheses for behavior prediction,} \emph{arXiv preprint arXiv:1910.05449}, 2019.

\bibitem[{Nakamura and Bansal(2023)}]{nakamura2023online}
Nakamura, K., and Bansal, S., \enquote{Online update of safety assurances using confidence-based predictions,} \emph{2023 IEEE International Conference on Robotics and Automation (ICRA)}, IEEE, 2023, pp. 12765--12771.

\bibitem[{Robb(2014)}]{robb2014system}
Robb, J., \enquote{System wide information management (SWIM): program overview and status update,} \emph{2014 Integrated Communications, Navigation and Surveillance Conference (ICNS) Conference Proceedings}, IEEE, 2014, pp. 1--15.

\bibitem[{Vinoski(2006)}]{vinoski2006advanced}
Vinoski, S., \enquote{Advanced message queuing protocol,} \emph{IEEE Internet Computing}, Vol.~10, No.~6, 2006, pp. 87--89.

\bibitem[{Glasmacher et~al.(2022)Glasmacher, Krajewski, and Eckstein}]{glasmacher2022automated}
Glasmacher, C., Krajewski, R., and Eckstein, L., \enquote{An automated analysis framework for trajectory datasets,} \emph{arXiv preprint arXiv:2202.07438}, 2022.

\bibitem[{Melnychuk et~al.(2022)Melnychuk, Frauen, and Feuerriegel}]{melnychuk2022causal}
Melnychuk, V., Frauen, D., and Feuerriegel, S., \enquote{Causal transformer for estimating counterfactual outcomes,} \emph{International Conference on Machine Learning}, PMLR, 2022, pp. 15293--15329.

\bibitem[{Vaswani et~al.(2017)Vaswani, Shazeer, Parmar, Uszkoreit, Jones, Gomez, Kaiser, and Polosukhin}]{vaswani2017attention}
Vaswani, A., Shazeer, N., Parmar, N., Uszkoreit, J., Jones, L., Gomez, A.~N., Kaiser, {\L}., and Polosukhin, I., \enquote{Attention is all you need,} \emph{Advances in neural information processing systems}, Vol.~30, 2017.

\end{thebibliography}
